\documentclass[letterpaper]{article} 
\usepackage{aaai24}  
\usepackage{times}  
\usepackage{helvet}  
\usepackage{courier}  
\usepackage[hyphens]{url}  
\usepackage{graphicx} 
\usepackage{natbib}  
\usepackage{caption} 
\usepackage{amsmath}
\usepackage{amssymb}
\usepackage[utf8]{inputenc}
\usepackage{booktabs}
\usepackage{amsfonts}
\usepackage{nicefrac} 
\usepackage{microtype} 
\usepackage{xcolor}
\usepackage{enumitem}
\usepackage{array}
\usepackage{amsthm}
\usepackage{multirow}
\usepackage{mathtools}
\usepackage{graphics}
\usepackage{caption,subcaption}
\usepackage{mathtools}

\DeclarePairedDelimiter\ceil{\lceil}{\rceil}
\DeclarePairedDelimiter\floor{\lfloor}{\rfloor}
\usepackage{algorithm,algpseudocode}
\algdef{SE}[SUBALG]{Indent}{EndIndent}{}{\algorithmicend\ }%
\algtext*{Indent}
\algtext*{EndIndent}

\theoremstyle{definition}

\newtheorem{definition}{Definition}[section]

\newtheorem{assumption}{Assumption}[section]

\newtheorem{theorem}{Theorem}[section]

\usepackage{newfloat}
\usepackage{listings}
\DeclareCaptionStyle{ruled}{labelfont=normalfont,labelsep=colon,strut=off} 
\floatstyle{ruled}
\newfloat{listing}{tb}{lst}{}
\floatname{listing}{Listing}
\title{Inconsistency-Based Data-Centric Active Open-Set Annotation}
\author{
    Ruiyu Mao, Ouyang Xu, Yunhui Guo
}
\affiliations{
   The University of Texas at Dallas\\
    \{ruiyu.mao, oxu, yunhui.guo\}@utdallas.edu

}

\usepackage{bibentry}

\begin{document}

\maketitle

\begin{abstract}
Active learning, a method to reduce labeling effort for training deep neural networks, is often limited by the assumption that all unlabeled data belong to known classes. This closed-world assumption fails in practical scenarios with unknown classes in the data, leading to active open-set annotation challenges. Existing methods struggle with this uncertainty. We introduce \textsc{Neat}, a novel, computationally efficient, data-centric active learning approach for open-set data. \textsc{Neat} differentiates and labels known classes from a mix of known and unknown classes, using a clusterability criterion and a consistency measure that detects inconsistencies between model predictions and feature distribution. In contrast to recent learning-centric solutions, \textsc{Neat} shows superior performance in active open-set annotation, as our experiments confirm. Our implementations for the metrics and datasets are publicly available at: {\textcolor{purple}{https://github.com/RuiyuM/Active-OpenSet-NEAT}}.

\end{abstract}

\section{Introduction}
The remarkable performance of modern deep neural networks owes much to the availability of large-scale datasets such as ImageNet \cite{deng2009imagenet}. However, creating such datasets is a challenging task that requires a significant amount of effort to annotate data points \cite{sorokin2008utility,snow2008cheap,raykar2010learning,welinder2010multidimensional}. Fortunately, active learning offers a solution by enabling us to label only the most \emph{significant} samples \cite{settles2009active,cohn1996active,settles2011theories,beygelzimer2009importance,gal2017deep}. In active learning, a small set of labeled samples from known classes is combined with unlabeled samples, with the goal of selecting specific samples to label from the pool to enhance model training. Typical active learning methods for training deep neural networks involve selecting samples that have high levels of uncertainty \cite{settles2009active,cohn1994improving, balcan2006agnostic}, are in close proximity to the classification boundary \cite{ducoffe2018adversarial}, or reveal cluster structures from the data
\cite{sener2017active,ash2019deep}.

Despite the considerable body of research on active learning, its practical implementation in an \emph{open-world} context remains relatively unexplored \cite{ning2022active}. Unlike in closed-world active learning settings, where the unlabeled data pool is assumed to consist only of known classes, the open world introduces unknown classes into the unlabeled data pool, making active open-set annotations a challenge. For example, when collecting data to train a classifier to distinguish between different dog breeds as known classes, a large number of unlabeled images collected from online sources may include images from unknown classes such as wolves and coyotes. The task is to identify images with known classes and select informative samples for labeling, while avoiding samples from unknown classes. Existing active learning methods face a significant obstacle in this task, as they may tend to select samples from unknown classes for labeling due to their high uncertainty \cite{ning2022active}, which is undesirable.
\begin{figure}[!t]
\includegraphics[width=0.999\linewidth]{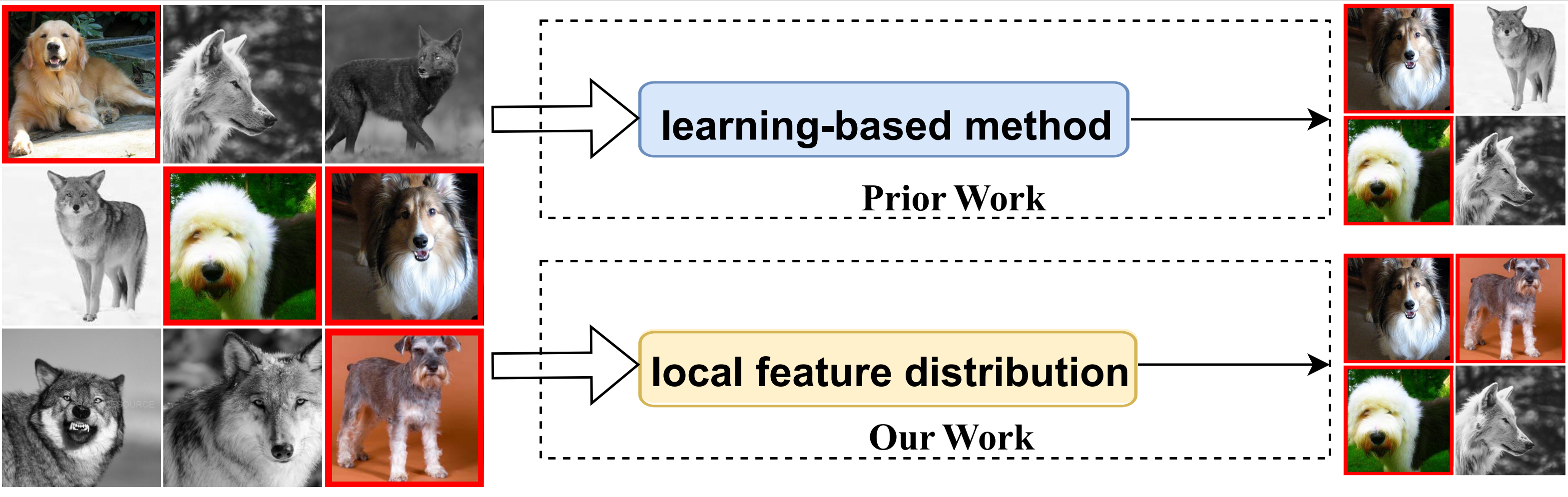}
\caption{Dataset consists of color images as known dog class and gray-scale images as unknown wolf class. Prior work using learning-based approach may identify some unknown classes as known classes. Our work focusing on local feature distribution can find known classes more accurately.
}
\label{fig:overview}
\end{figure}

Addressing the active open-set annotation problem requires specialized methods, and one such approach, called \textsc{LfOSA} was recently proposed by \citet{ning2022active}. \textsc{LfOSA} is a learning-based active open-set annotation as it involves training a detector network with an additional output for unknown classes. The predictions of the detector network on unlabeled data are used for identifying known classes. While this approach can achieve impressive performance, it has two limitations: 1) training the additional detector network is costly, 2) and it is difficult to identify informative samples from the known classes as there is a contradiction in that while excluding unknown classes is necessary, it is also easy to exclude informative samples from the known classes.

In this paper, we propose a novel inconsistency-based data-centric active learning method to actively annotate informative samples in an open world, which not only reduces the computational cost but also improves the performance of active open set annotation (Figure \ref{fig:overview}). Rather than using a learning-based approach to differentiate known and unknown classes, we suggest a data-centric perspective that naturally separates them by label clusterability, eliminating the need for an additional detector network. In addition, our method involves selecting informative samples from known classes by estimating the inconsistency between the model prediction and the local feature distribution. For example, suppose the model predicts an unlabeled sample as a wolf, but the majority of nearby samples are actually dogs. In that case, we would choose to label this unlabeled sample. The proposed inconsistency-based active learning approach shares a similar spirit to the version-space-based approach \cite{cohn1994improving, balcan2006agnostic, dasgupta2005coarse}. However, there are two key differences. Firstly, our hypothesis class consists of deep neural networks. Secondly, the version-space-based approach identifies uncertain examples by analyzing the consistencies among multiple models, our approach leverages a fixed model and estimates the consistencies between the model prediction and the local feature distribution.

The contributions of this paper are summarized below,

\begin{itemize}
    \item We introduce \textsc{Neat}, a novel and efficient inconsistency-based data-centric active learning method. This method is designed to select informative samples from known classes within a pool containing both known and unknown classes. To the best of our knowledge, our work stands as the pioneering effort in utilizing large language models for active open-set annotation.

    \item Compared with the learning-based method, the proposed data-centric active learning method is more computationally efficient and can effectively identify informative samples from the known classes.
    \item Extensive experiments show that \textsc{Neat} achieves much better results compared to standard active learning methods and the method specifically designed for active open set annotation. In particular, \textsc{Neat} achieves an average accuracy improvement of 9\% on \textbf{CIFAR10}, \textbf{CIFAR100} and \textbf{Tiny-ImageNet} compared with existing active open-set annotation method given the same labeling budget.
\end{itemize}

\section{Related Work}
Active learning, a topic which is extensively studied in machine learning \cite{lewis1995sequential,settles2009active,cohn1996active,settles2011theories,beygelzimer2009importance}, focuses on selecting the most uncertain samples for labeling. Uncertainty-based active learning methods \cite{lewis1995sequential} employ measures such as entropy \cite{shannon1948mathematical}, least confident \cite{settles2009active}, and margin sampling \cite{scheffer2001active}. Another widely used approach is the version-space-based method \cite{cohn1994improving, balcan2006agnostic, dasgupta2005coarse, dasgupta2011two} which maintains multiple models consistent with labeled samples. If these models make inconsistent predictions on an unlabeled sample, a label query is prompted.

For deep neural networks, specific active learning methods have been developed. For instance, Core-Set \cite{sener2017active} advocates for an active learning method that selects samples which are representative of the whole data distribution. \textsc{Badge}  \cite{ash2019deep} selects samples based on predictive uncertainty and sample diversity. In particular, \textsc{Badge}  leverages \textsc{$k$-Means++} to select a set of samples which have diverse gradient magnitudes. The approach \textsc{BGADL} \cite{tran2019bayesian} integrates active learning and data augmentation. It leverages Bayesian inference in the generative model to create new training samples from selected existing samples. These generated samples are then utilized to enhance the classification accuracy of deep neural networks. \textsc{CEAL} \cite{wang2016cost} combines pseudo-labeling of clearly classified samples and actively labeled informative samples to train deep neural networks. For deep object detection, active learning is introduced in \cite{brust2018active}, utilizing prediction margin to identify valuable instances for labeling. \textsc{DFAL} \cite{ducoffe2018adversarial} is an adversarial active learning method for deep neural networks that selects samples based on their distance to the decision boundary, approximated using adversarial examples. The samples closest to the boundary are chosen for classifier training. Recently, \textsc{LfOSA} \cite{ning2022active} is proposed as the first active learning method for active open-set annotation. \textsc{LfOSA} trains a detector network to identify known classes and selects samples based on model confidence. \textsc{MQNet} \cite{park2022metaquerynet} is an open-set active learning method that addresses the Purity-Informativeness Dilemma and dynamically solves it using a  Meta-learning MLP (Multilayer Perceptron) network, based on the samples' purity and informativeness scores.

\section{Preliminary}

\subsection{Active Open-Set Annotation}

Assume the sample-label pair $(X, Y)$ has the joint distribution $P_{XY}$. For a given sample $\mathbf{x}$, the label $y$ can be determined via the conditional expectation,
\begin{equation}
    \eta(\mathbf{x}) = \mathbb{E}[Y|X = \mathbf{x}]
\end{equation}
We consider a pool-based active learning setup. We begin by randomly sampling a small labeled set denoted as $L = \{\mathbf{x}_i, y_i \}_{i=1}^N$ from the joint distribution $P_{XY}$. We also have access to a large set of unlabeled samples, denoted as $U$. Given a deep neural network $f_\theta$ parameterized by $\theta$, the expected risk of the network is computed as $R = \mathbb{E}[\ell(y, f_\theta(\mathbf{x}))] = \int l(y, f_\theta(\mathbf{x}))d P_{XY}$, where $\ell(y, f_\theta(\mathbf{x}))$ represents the classification error. We conduct $T$ query rounds, where in each round $t$, we are allotted a fixed labeling budget $B$ to identify $B$ informative samples from $U$, denoted as $U_B$. The queried samples $U_B$ are given labels and added to the initial labeled set, with the aim of minimizing the sum of cross-entropy losses $\sum_{(\mathbf{x}, y) \in L \bigcup U_B} \ell_{\textnormal{CE}}(f_\theta(\textbf{x}), y)$ and reducing the expected risk of the model $f_\theta$ \cite{settles2009active}. The term $\ell_{\textnormal{CE}}(f_\theta(\mathbf{x}), y)$ represents the cross-entropy loss between the predicted and true labels for a sample $(\mathbf{x}, y)$ \cite{bishop2006pattern}.

In the context of active open-set annotation \cite{ning2022active}, the set of unlabeled samples $U$ comprises data from both known classes and unknown classes, denoted as $S_{known}$ and $S_{unknown}$, respectively. Importantly, $S_{known}$ represents the classes observed during the initial labeling process, and it is ensured that $S_{known} \cap S_{unknown} = \emptyset$. In each active query round, we label the samples from known classes with their true labels and label the samples from unknown classes as \emph{invalid}. Active open-set annotation presents a greater challenge than standard active learning, as it requires the active learning method { \bf 1)} to distinguish between known and unknown classes among the unlabeled samples { \bf 2)} and to select informative samples exclusively from the known classes.

\subsection{Learning-Based Active Open-Set Annotation}
The study of active open-set annotation is currently limited. A recent paper \cite{ning2022active} proposed a new learning-based method called \textsc{LfOSA} which is specially designed for active open-set annotation. This approach combines a model, $f_\theta$, trained on known classes with a detector network that identifies unknown class samples. Suppose there are $C$ known classes, the detector network has $C+1$ outputs, similar to \textsc{OpenMax}\cite{bendale2016towards}, which allow for the classification of unlabeled data into known and unknown classes. During an active query, \textsc{LfOSA} focuses only on unlabeled data classified as known classes, and then uses a Gaussian Mixture Model to cluster the activation values of detected known classes data into two clusters. The data closer to the larger cluster mean is selected for labeling. This learning-based approach has shown significant improvement over traditional active learning methods for active open-set annotation. However, there are still challenges that need to be addressed, such as the additional computation cost required to train the detector network and the difficulty of identifying informative samples from the known classes.

\begin{algorithm}[!t]
\small
\caption{\textsc{Neat}: Inconsistency-Based Data-Centric Active Open-Set annotation.}
\label{alg:neat}
\begin{algorithmic}[1]
 \Require A deep neural network $f_\theta$, initial labeled set $L$, a set of known class $Y_\textnormal{Known}$  , unlabeled labeled set $U$, number of query rounds $T$, number of examples in each query batch $B$, a pre-trained model $M$, number of neighbors $K$.
\State Use the pre-trained model $M$ for extracting features on $L$ and $U$.
\For{$t \gets 1$ to $T$}
    \State Train the model $f_\theta$ on $L$ by minimizing $\sum_{(x, y) \in L} \ell_{\textnormal{CE}}(f_\theta(x), y)$
    \State $S \gets \{\}$.
    \State For each sample $x \in U$:
        \Indent
            \State Compute the output of the softmax function as $P_x$.
            \State Find the $K$-nearest neighbors $\{ N_i(\mathbf{x})\}_{i=1}^K$ of $\mathbf{x}$ in $L$ based on the extracted features.
            \State If all the labels from $\{ N_i(\mathbf{x})\}_{i=1}^K$ belong to known classes $Y_\textnormal{Known}$, then $S \gets S \bigcup \{\mathbf{x}\}$. 
        \EndIndent

    \State Compute the score $I(x)$ using Eq. \ref{eq: score} for each sample $\mathbf{x} \in S$.
    \State Rank the samples based on $I(\mathbf{x})$ and denote the B samples which have the largest scores as $U_B$.
    \State Query the labels of each sample in $U_B$.
    \State $U \gets U \setminus U_B$, $L \gets L\bigcup U_B$.
\EndFor
\end{algorithmic}
\end{algorithm}

\section{\textsc{Neat}}
Algorithm \ref{alg:neat} describes the detailed algorithm of \textsc{Neat}. Initially, \textsc{Neat} begins with a randomly drawn labeled set $L$ from known classes. In each query round, \textsc{Neat} performs two main steps. First, it identifies unlabeled samples whose neighbors in the labeled set belong to known classes. Secondly, \textsc{Neat} selects a batch of unlabeled samples for labeling by estimating the inconsistency between the model's prediction and the local feature distribution. Unlike the learning-based method discussed in \cite{ning2022active}, \textsc{Neat} is more computationally efficient. Furthermore, it effectively decouples the detection of known classes from the identification of informative samples, enabling it to identify informative samples even from the known classes, which is a challenge for existing active learning methods.
 
\subsection{Data-Centric Known Class Detection}
In the learning-based method \cite{ning2022active}, an additional detector network is trained to differentiate known and unknown classes. However, the training cost is high. In contrast, \textsc{Neat} adopts a data-centric perspective for finding known classes samples based on feature similarity. In particular, \textsc{Neat} relies on label clusterability to identify known classes from the unlabeled pool.

\noindent\textbf{Label clusterability.} We leverage the intuition that samples with similar features should belong to the same class \cite{gao2016resistance,zhu2021clusterability,zhu2022detecting}. This intuition can be formally defined as follows,

\begin{definition} (($K$, $\sigma_K$) label clusterability).
A dataset $D$ satisfies $(K, \sigma_K)$ label clusterability if for all $\mathbf{x} \in D$, the sample $\mathbf{x}$ and its $K$-Nearest-Neighbors ($K$-NN) $\{\mathbf{x}_1, \mathbf{x}_2, ..., \mathbf{x}_K\}$ belong to the same label with probability at least $1 - \sigma_K$.
\label{def: label}
\end{definition}
\noindent  When $\sigma_K$ = 0, then it is called $K$-NN label clusterability \cite{zhu2021clusterability}. Recent methods for noisy label learning \cite{zhu2021clusterability,zhu2022detecting} leverage label clusterability for detecting examples with noisy labels. 

\noindent\textbf{Feature extraction.} To utilize the clusterability of labels for identifying known classes, we need to extract features from unlabeled inputs that group semantically similar samples in the feature space. Additionally, the quality of these features will inevitably impact the clusterability of the labels. Instead of developing a separate detector, as demonstrated in \citet{ning2022active}, we suggest leveraging pre-trained large language models for feature extraction, which have been demonstrated to possess exceptional zero-shot learning ability \cite{radford2021learning}. In particular, we leverage CLIP \cite{radford2021learning} to extract features for both the labeled and unlabeled data, providing high-quality features for calculating feature similarity.

\noindent\textbf{Known class detection. } By utilizing the features extracted by CLIP from both the labeled set $L$ and the unlabeled set $U$, we can identify the $K$-nearest neighbors $\{ N_1(\mathbf{x}), N_2(\mathbf{x}), ..., N_K(\mathbf{x}) \}$ in the labeled set $L$ for each sample $\mathbf{x} \in U$, using cosine distance. Each $N_k(\mathbf{x}) \in L$ represents the $k$-th closest samples in $L$ to the unlabeled sample $\mathbf{x}$. Subsequently, we compute the count of neighbors with known and unknown classes for each unlabeled sample $\mathbf{x}$, assuming label clusterability. If an unlabeled sample is in proximity to numerous known class samples, it is more likely to belong to a known class. Therefore, we classify an unlabeled sample as belonging to a known class if all of its neighboring samples are from known classes.

\noindent\textbf{Theoretical analysis}. Here we give an upper bound on the detection error of the proposed known class detection. The first step is to understand in what condition our method will make a mistake. Given the sample $\mathbf{x}$, although all of its $K$-nearest neighbors $\{ N_1(\mathbf{x}), N_2(\mathbf{x}), ..., N_K(\mathbf{x})\}$ belong to known classes, it is still possible that the sample is from the unknown classes. This can caused by the randomness in the labeling process and the quality of the features. Instead of using the definition \ref{def: label} which is too coarse for our analysis, we introduce the following assumptions,
\begin{assumption}
There exists a constant $C > 0$ and $0 < \alpha < 1$ such that for any $\mathbf{x}$ and  $\mathbf{x}'$,
\begin{equation} 
    \mathbb{P}( y_{true}(\mathbf{x}) \neq y_{true}(\mathbf{x}')) \le C \rho(\mathbf{x}, \mathbf{x}')^\alpha
    \label{assump: dist1}
\end{equation}
\end{assumption}

\noindent Here, $\rho(\mathbf{x}, \mathbf{x}')$ represents the distance between samples $\mathbf{x}$ and $\mathbf{x}'$, while $y_{true}(\mathbf{x})$ and $y_{true}(\mathbf{x}')$ correspond to the true labels of $\mathbf{x}$ and $\mathbf{x}'$, respectively. We use $r_K(\mathbf{x})$ to denote the radius of the ball centered at $\mathbf{x}$ such that $\forall$ $\mathbf{x}'$ in the $K$-nearest neighbors of $\mathbf{x}$, $\rho(\mathbf{x}, \mathbf{x}') \le r_K(\mathbf{x})$. 

\begin{assumption}
For any $\mathbf{x}$, the labeling error is upper-bounded by a small constant $e$,
\begin{equation} 
    \mathbb{P}( y(\mathbf{x}) \neq y_{true}(\mathbf{x})) \le e
    \label{assump: dist2}
\end{equation}

\end{assumption}

Then, we can establish an upper bound for the detection error as follows,
\begin{theorem} Given the assumption 0.1. and 0.2. and the number of neighbors $K$, the probability of making a detection error is upper-bounded as,
\begin{align}
\begin{split}
\mathbb{P}(\textnormal{Error} | & K) \le  \sum_{k=\ceil{\frac{K - 1}{2}} + 1}^K \binom Kk e^k(1-e)^{K-k}  C^k r_K(\mathbf{x})^{\alpha k} \\
& + \sum_{k=0}^{  \floor{\frac{K+1}{2}}-1 } \binom Kk e^k(1-e)^{K-k}  C^{(K-k)} r_K(\mathbf{x})^{\alpha (K-k)}
\end{split}
\end{align}

\end{theorem}

The proof of the theorem will be given in the Appendix. From the theorem, we can conclude that a good feature will decrease the detection error as we expected.

\subsection{Inconsistency-Based Active Learning}
To improve accuracy with a fixed labeling budget, it is crucial to select informative samples for labeling. One active learning strategy, motivated by theory, is the version-space-based approach \cite{cohn1994improving, balcan2006agnostic, dasgupta2005coarse, dasgupta2011two}. This approach involves maintaining a set of models that are consistent with the current labeled data, and an unlabeled sample is selected for labeling if two models produce different predictions. However, implementing this approach for deep neural networks is challenging due to the computational cost of training multiple models \cite{ash2019deep}. To address this issue, we propose an inconsistency-based active learning method that does not require training multiple models and naturally leverages features produced by CLIP.

Given an unlabeled sample $\mathbf{x} \in U$, the model $f_\theta$'s prediction for $\mathbf{x}$, denoted as $P_\mathbf{x} \in \mathbb{R}^{C}$, represents a probability vector where $P_\mathbf{x}[c]$ is the model's confidence that $\mathbf{x}$ belongs to class $c$. Since we lack ground-truth labels, measuring prediction accuracy is impossible. To address this, we propose evaluating the importance of the sample for improving model training by assessing whether the model prediction aligns with local feature similarity. For example, if the model predicts an unlabeled sample as a dog with low probability but the majority of the sample's neighbors belong to the dog class, then either the model's prediction is incorrect or the sample is near the decision boundary between the dog class and the true class. In either case, the sample can be labeled to improve model training.

Given the $K$-nearest neighbors $\{ N_i(\mathbf{x})\}_{i=1}^K$ of the example $\mathbf{x}$, we first construct a vector $V_\mathbf{x} \in \mathbb{R}^{C}$ with $V_\mathbf{x}[c] = \sum_k \mathbf{1}( Y_k(\mathbf{x}) = c )$, where $Y_k(\mathbf{x}) $ is the label of $k$-th nearest neighbor of $\mathbf{x}$ and $\mathbf{1}(\cdot)$ is an indicator function. Then the vector $V_\mathbf{x} $ is normalized via the softmax function to be a probabilistic vector $\Tilde{V}_\mathbf{x}$. The inconsistency between the model prediction and local feature similarity is computed using cross-entropy as,
\begin{equation}
    I(\mathbf{x}) = -\sum_{c=1}^C P_\mathbf{x}[c] \log \Tilde{V}_\mathbf{x}[c].
    \label{eq: score}
\end{equation}
A large $I(\mathbf{x})$ indicates that the model prediction is inconsistent with local feature distribution. Similar to version-space-based approach, the unlabeled sample is selected for labeling. In each query round, we rank all the identified known classes samples in the first stage using $I(\mathbf{x})$ and select the top $B$ samples for labeling.

\section{Experiments}

\subsection{Experimental Settings and Evaluation Protocol}

\noindent\textbf{Datasets and models.} We consider \textbf{CIFAR10} \cite{krizhevsky2009learning}, \textbf{CIFAR100} \cite{krizhevsky2009learning}, and \textbf{Tiny-Imagenet} \cite{le2015tiny}, to evaluate the performance of our proposed method. Similar to the existing methods for active open-set annotations \cite{ning2022active}, we leverage a ResNet-18 \cite{he2016deep} architecture to train the classifier for the known classes. For the proposed method, we leverage CLIP \cite{radford2021learning} to extract the features for both known and unknown classes.

\noindent\textbf{Active open-set annotation.} In accordance with \cite{ning2022active}, the experiment randomly selects 40 classes, 20 classes and 2 classes from \textbf{Tiny-Imagenet}, \textbf{CIFAR100}, and \textbf{CIFAR10}, respectively, as known classes, while the remaining classes are treated as unknown classes. To begin with, following \cite{ning2022active}, 8\% of the data from the known classes in \textbf{Tiny-ImageNet} and \textbf{CIFAR100}, and 1\% of the data from the known classes in \textbf{CIFAR10} are randomly selected to form an initial labeled set. The rest of the known class data and the unknown class data are combined to form the unlabeled data pool.

\noindent\textbf{Baseline methods.} We consider the following active learning methods as baselines,
\begin{enumerate}[noitemsep,topsep=0pt,parsep=1pt,partopsep=0pt]
    \item \textsc{Random}: The naive baseline which randomly selects samples for annotation;
    \item \textsc{Uncertainty} \cite{settles2009active}: A commonly used active learning method which selects samples with the highest degree of uncertainty as measured by entropy;
    \item \textsc{Certainty} \cite{settles2009active}: A common baseline for active learning which selects samples with the highest degree of certainty as measured by entropy;
    \item \textsc{Coreset} \cite{sener2017active}: This method identifies a compact, representative subset of training data for annotation;
    \item \textsc{Bgadl} \cite{tran2019bayesian}: A Bayesian active learning method which leverages generative to select informative samples;
    \item \textsc{OpenMax} \cite{bendale2016towards}: A representative open-set classification method which can differentiate between known classes and unknown classes;
    \item \textsc{Badge} \cite{ash2019deep}: An active learning method designed for deep neural networks which select a batch of samples with diverse gradient magnitudes;
    \item \textsc{MQNet} \cite{park2022metaquerynet}:
    An open-set active learning method employs a Meta-learning MLP network to dynamically select samples based on their purity and informativeness scores.
    \item \textsc{LfOSA} \cite{ning2022active}: A learning-based active open-set annotation method which selects samples based on maximum activation value (MAV) modeled by a Gaussian Mixture Model;
    \item \textsc{Neat} (Passive): We also consider a baseline that is the passive version of \textsc{Neat}. In \textsc{Neat} (Passive), we do not leverage the proposed inconsistency-based active learning methods for selecting samples for labeling but instead randomly sample from the identified known classes samples.
\end{enumerate}
\begin{figure*}[!t]
    \centering
    \begin{tabular}{ccc}
        \includegraphics[width=5cm]{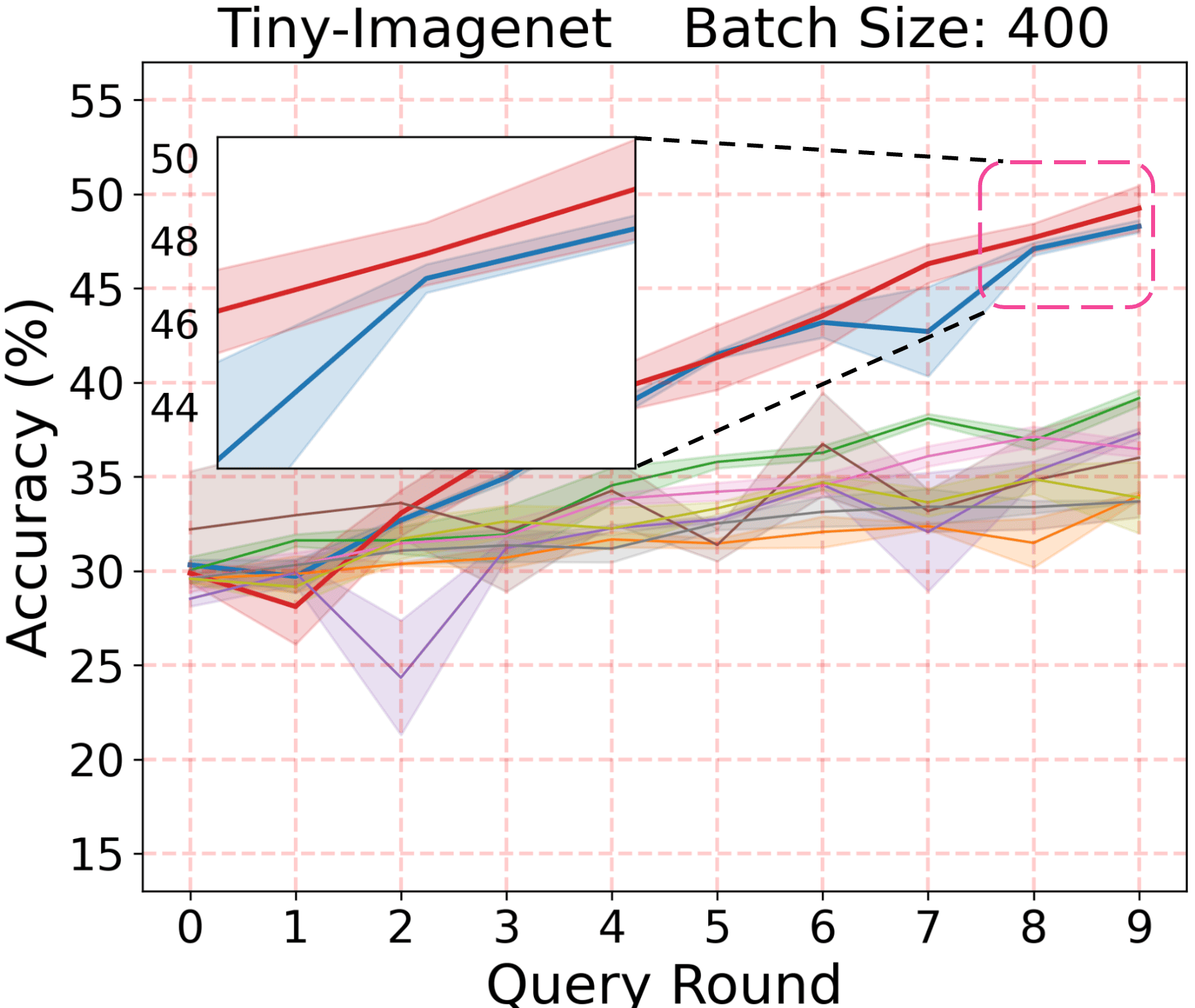} &
        \includegraphics[width=5cm]{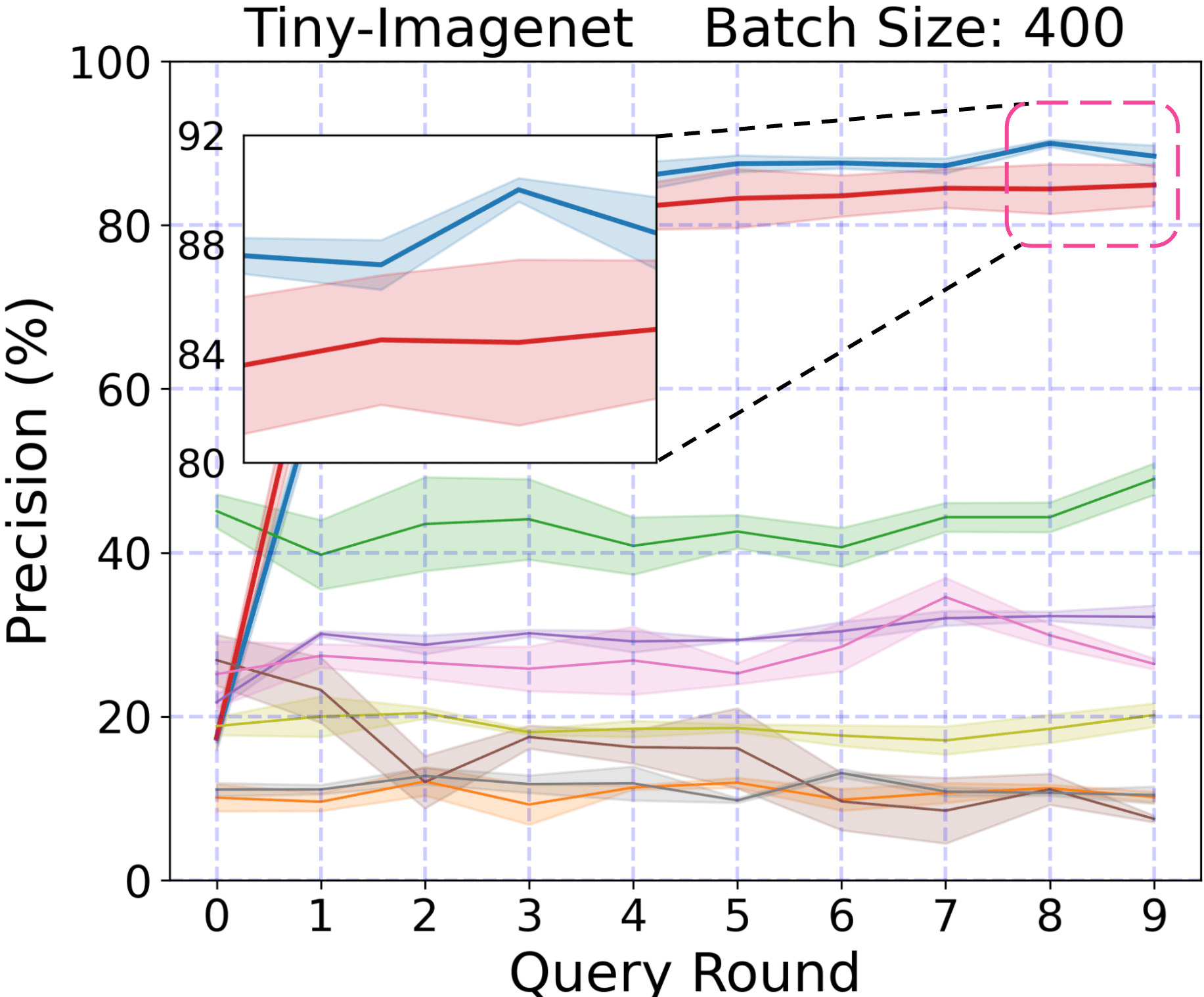} &
        \includegraphics[width=5cm]{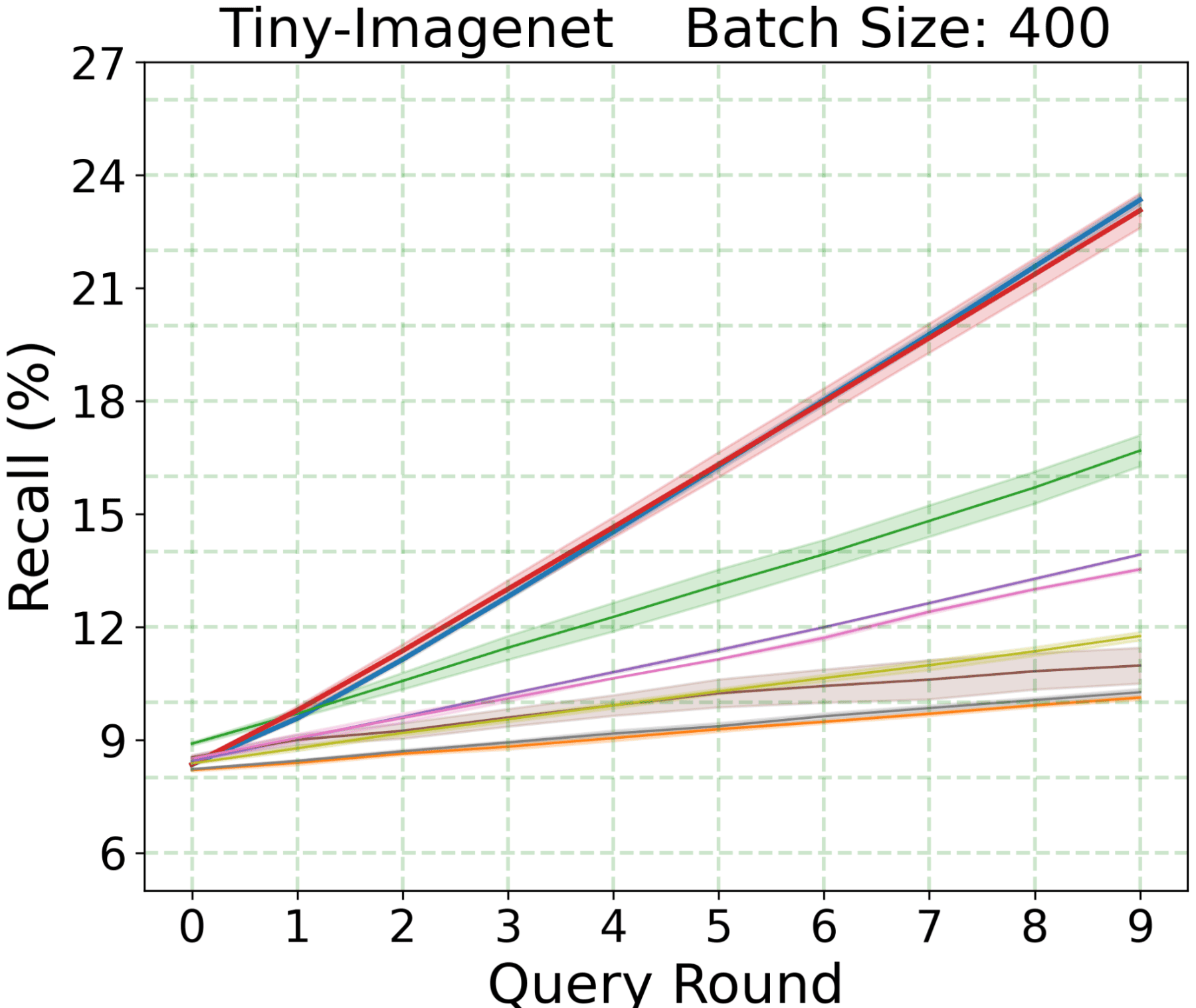} \\
        \includegraphics[width=5cm]{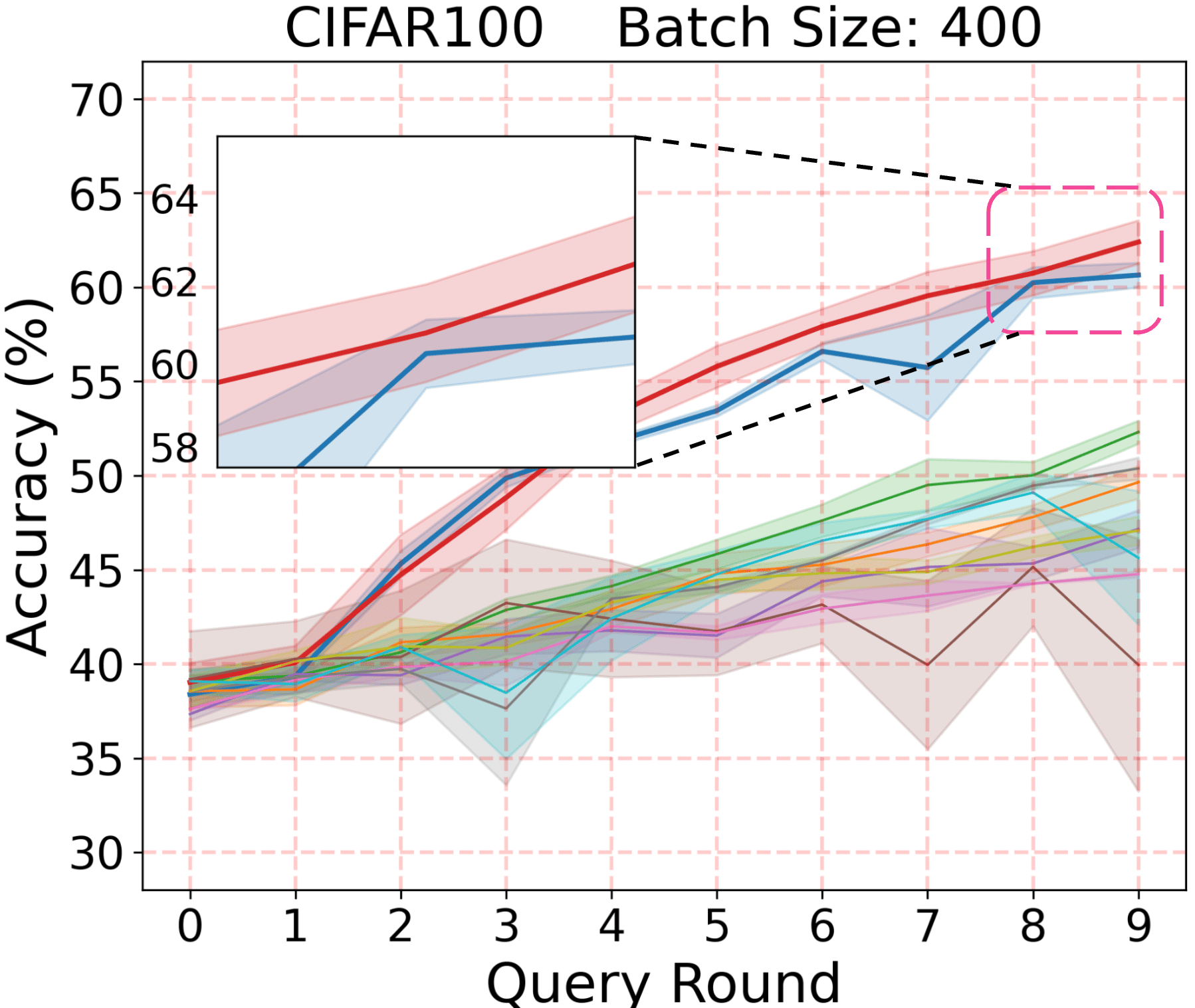} &
        \includegraphics[width=5cm]{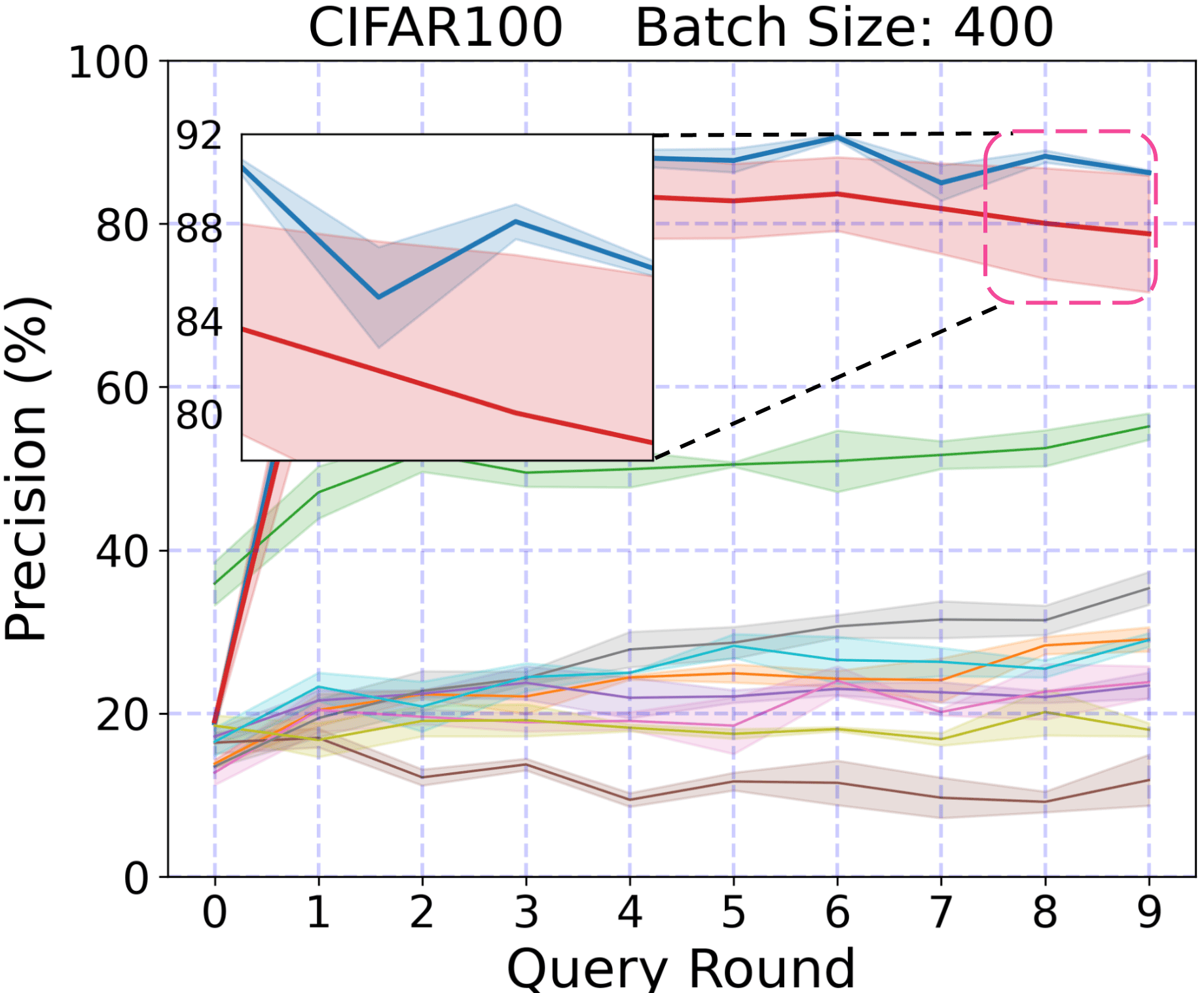} &
        \includegraphics[width=5cm]{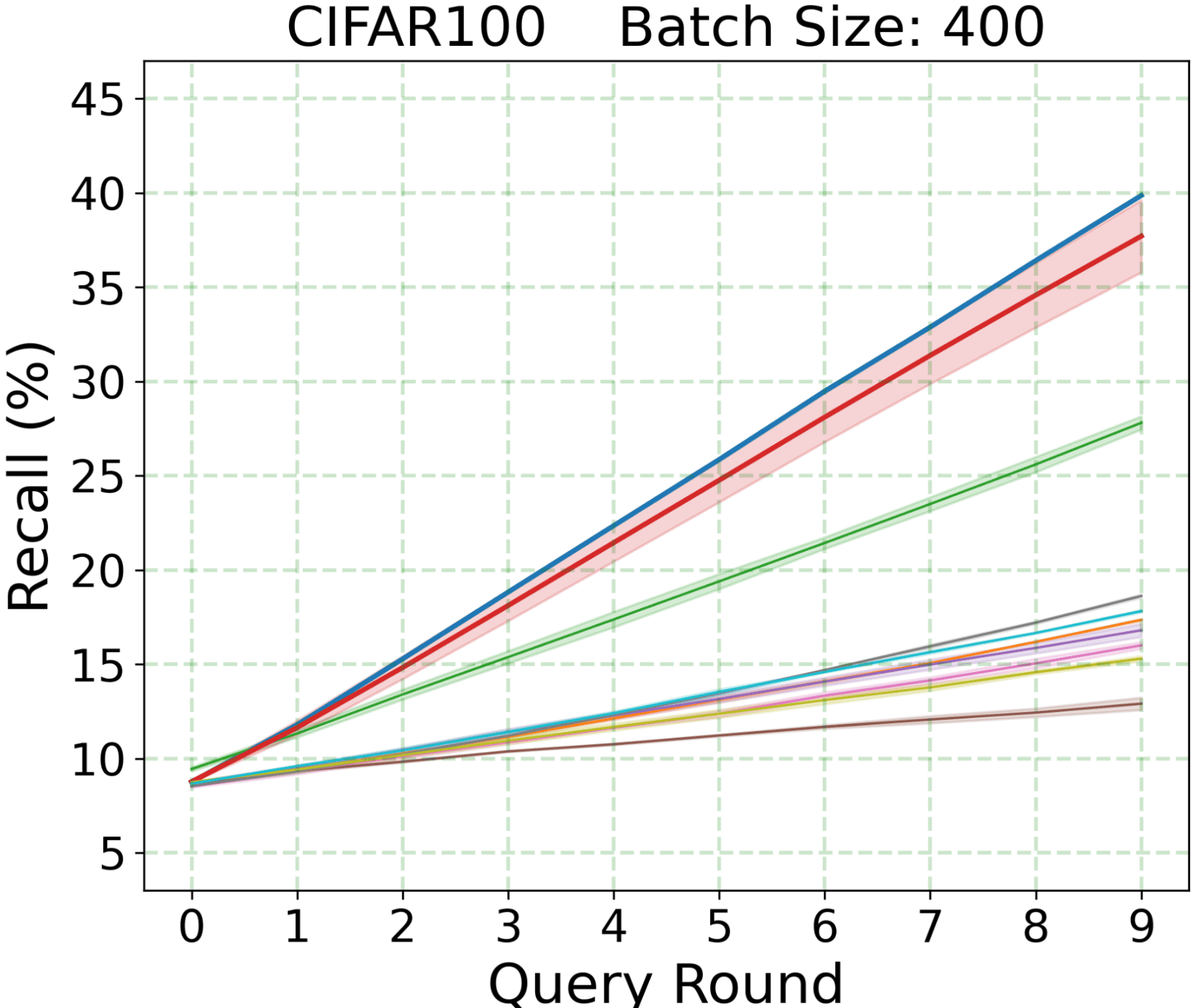} \\    
        \includegraphics[width=5cm]{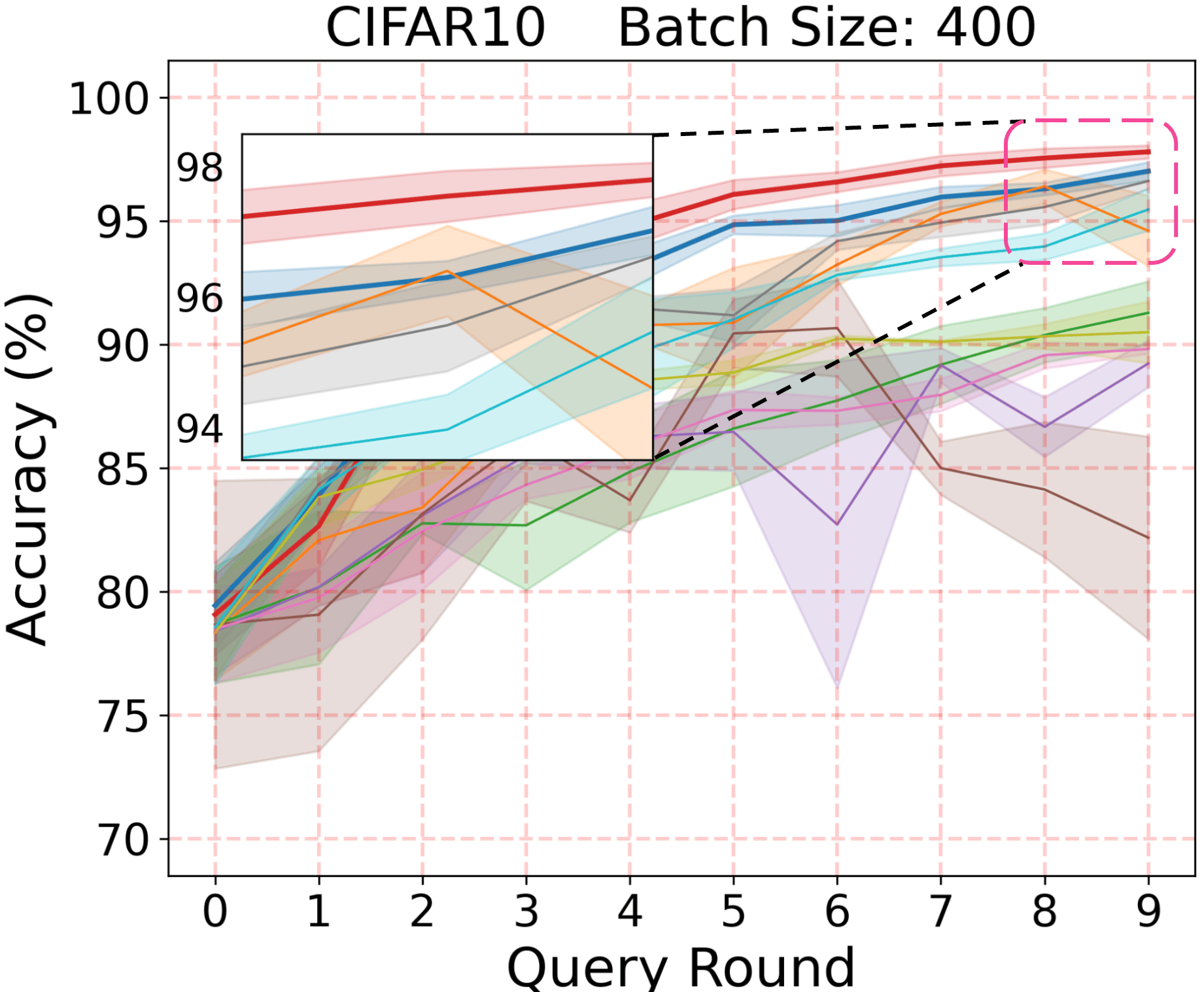} &
        \includegraphics[width=5cm]{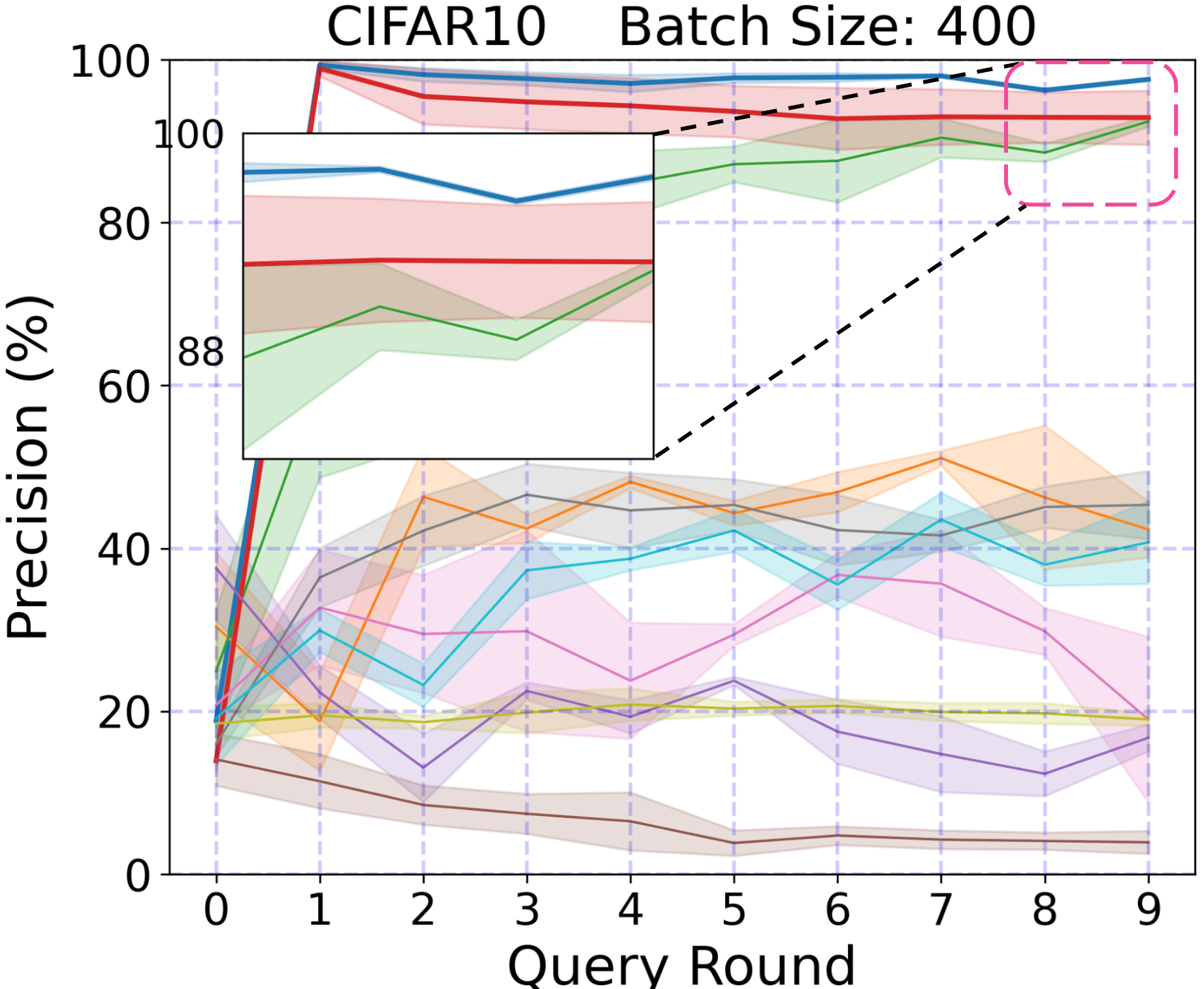} &
        \includegraphics[width=5cm]{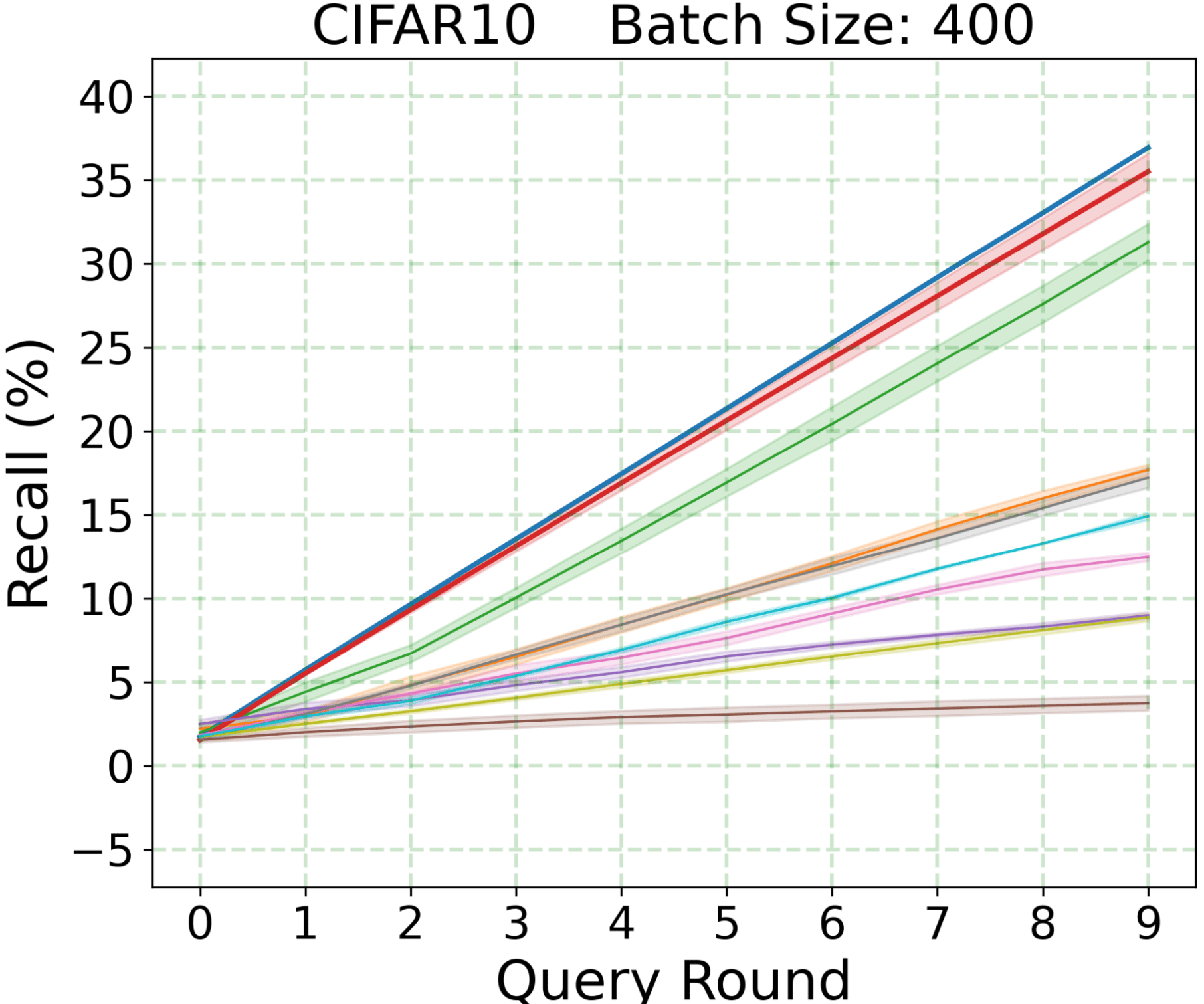} \\
        \multicolumn{3}{c}{\includegraphics[width=10cm]{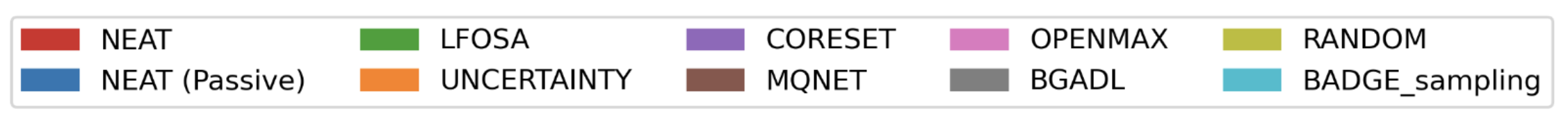}} \\              
    \end{tabular}
    \caption{\textsc{Neat} achieves higher precision, recall and accuracy compared with existing active learning methods for active open-set annotation. We evaluated \textsc{Neat} and the baseline active learning methods on \textbf{CIFAR10}, \textbf{CIFAR100} and \textbf{Tiny-ImageNet} based on accuracy, precision and recall. }
    \label{fig:baseline}
\end{figure*}
\noindent\textbf{Metrics.} We evaluate various active learning methods based on three key metrics: accuracy, precision, and recall. Accuracy reflects how accurately the model makes predictions on the test set. To measure recall, we use the ratio of selected known class samples to the total number of known class samples present in the unlabeled pool, denoted as $N^t_{\textnormal{known}}$ and $N^{\textnormal{total}}_{\textnormal{known}}$, respectively, in the query round $t$. Precision, on the other hand, measures the proportion of true known class samples among the selected samples in each query round,
\begin{equation}
    \textnormal{precision} = \frac{N^t_{\textnormal{known}}}{B} \quad\quad \textnormal{recall} = \frac{\sum_{t=1}^T N^t_{\textnormal{known}}}{N^{\textnormal{total}}_{\textnormal{known}}}
\end{equation}
\noindent\textbf{Implementations details.} The classification model, specifically ResNet-18,  was trained for 100 epochs using Stochastic Gradient Descent (SGD) with an initial learning rate of 0.01. The learning rate decayed by 0.5 every 20 epochs, and the training batch size is 128. There were 9 query rounds with a query batch size of 400. All experiments were repeated three times with different random seeds and the average results and standard deviations were reported. The experiments were conducted on four A5000 NVIDIA GPUs.

\subsection{Results}


\subsubsection{\textsc{Neat} VS. Baselines} We evaluate the performance of \textsc{Neat} and other methods by plotting curves as the number of queries increases (Figure \ref{fig:baseline}). It is evident that regardless of the datasets, \textsc{Neat} consistently surpasses other methods in all cases. In particular, \textsc{Neat} achieves much higher selection recall and precision compared to existing active learning methods which demonstrates the effectiveness of the data-centric known class detection. {\bf 1)} In terms of recall, \textsc{Neat} consistently outperforms other active learning methods by a significant margin. Notably, on \textbf{CIFAR10}, \textbf{CIFAR100}, and \textbf{Tiny-ImageNet}, \textsc{Neat} achieves improvements of 4\%, 12\%, and 7\%, respectively, compared to \textsc{LfOSA} \cite{ning2022active}, which is a learning-based active open-set annotation method. {\bf 2)} In terms of precision, \textsc{Neat} consistently maintains a higher selection precision than other baselines, with a noticeable gap. Importantly, \textsc{Neat}'s ability to differentiate between known and unknown classes is significantly improved by adding examples from unknown classes in the first query round. {\bf 3)} In terms of accuracy, \textsc{Neat} consistently outperforms \textsc{LfOSA} \cite{ning2022active} across all datasets, including \textbf{CIFAR10}, \textbf{CIFAR100}, and \textbf{Tiny-ImageNet}, with improvements of 6\%, 11\%, and 10\%, respectively. These results indicate that the proposed \textsc{Neat} method effectively addresses the open-set annotation (OSA) problem. It is worth noting that \textsc{Neat} (Passive) achieves slightly higher recall and precision than \textsc{Neat}. The possible reason is that by selecting informative samples from known classes, it is also more likely to include samples from unknown classes with high uncertainty. Despite having a slightly lower recall, \textsc{Neat} achieves a higher accuracy in all data sets compared to \textsc{Neat} (Passive). This demonstrates that \textsc{Neat} can effectively identify informative samples from known classes to train the model.
\subsubsection{The Use of Pre-Trained Models for Active Open-Set Annotation}
Unlike the majority of existing active learning methods, \textsc{Neat} incorporates a pre-trained model as an component of the method. Although there was some initial works that utilized large language models for natural language processing (NLP) tasks \cite{bai2020pre,seo2022active}, to our knowledge, our work represents the first instance of leveraging a large language model for active open-set annotation.

\begin{figure*}[!t]
    \centering
    \begin{tabular}{ccc}
        \includegraphics[width=0.26\textwidth]{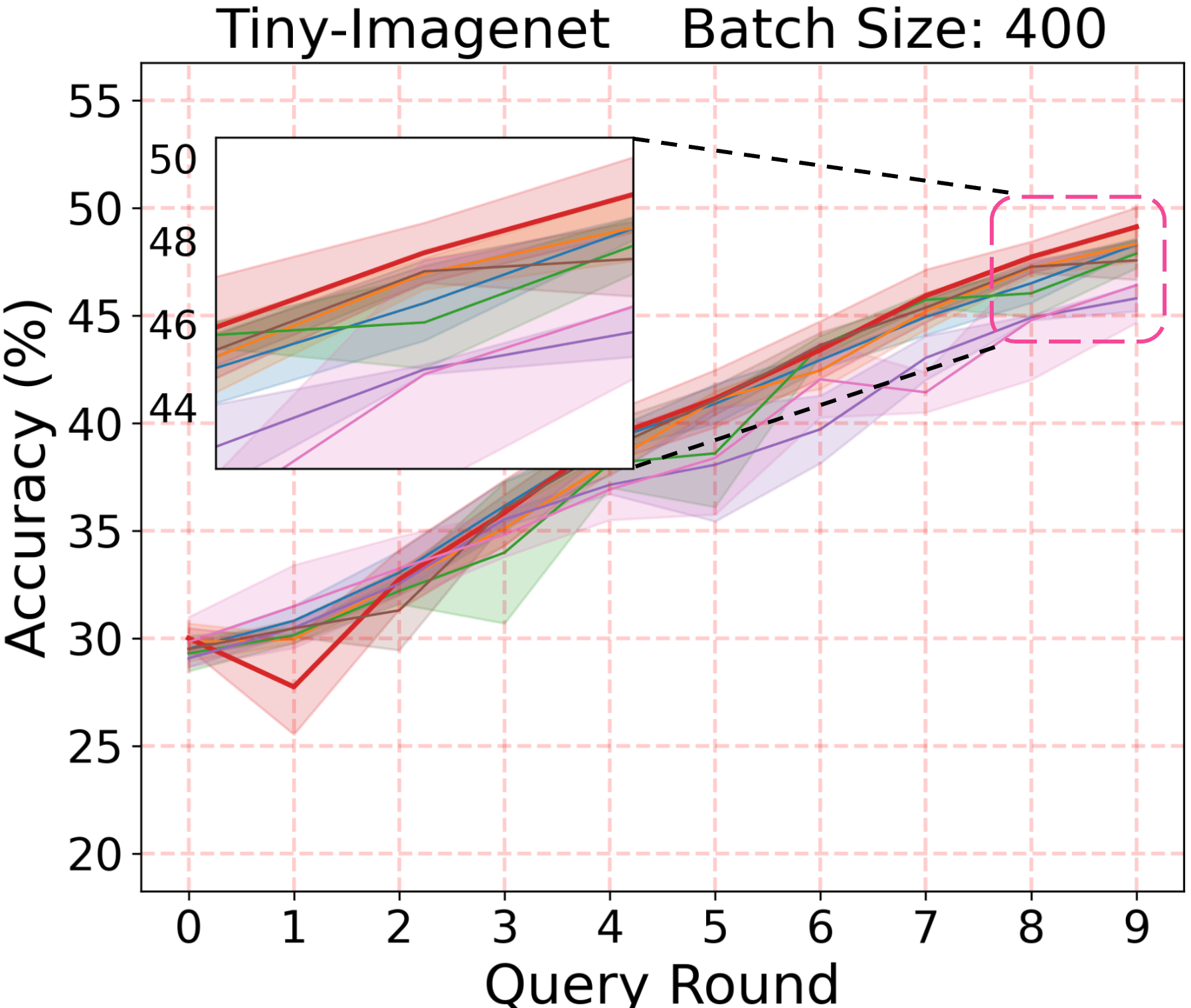} &
        \includegraphics[width=0.26\textwidth]{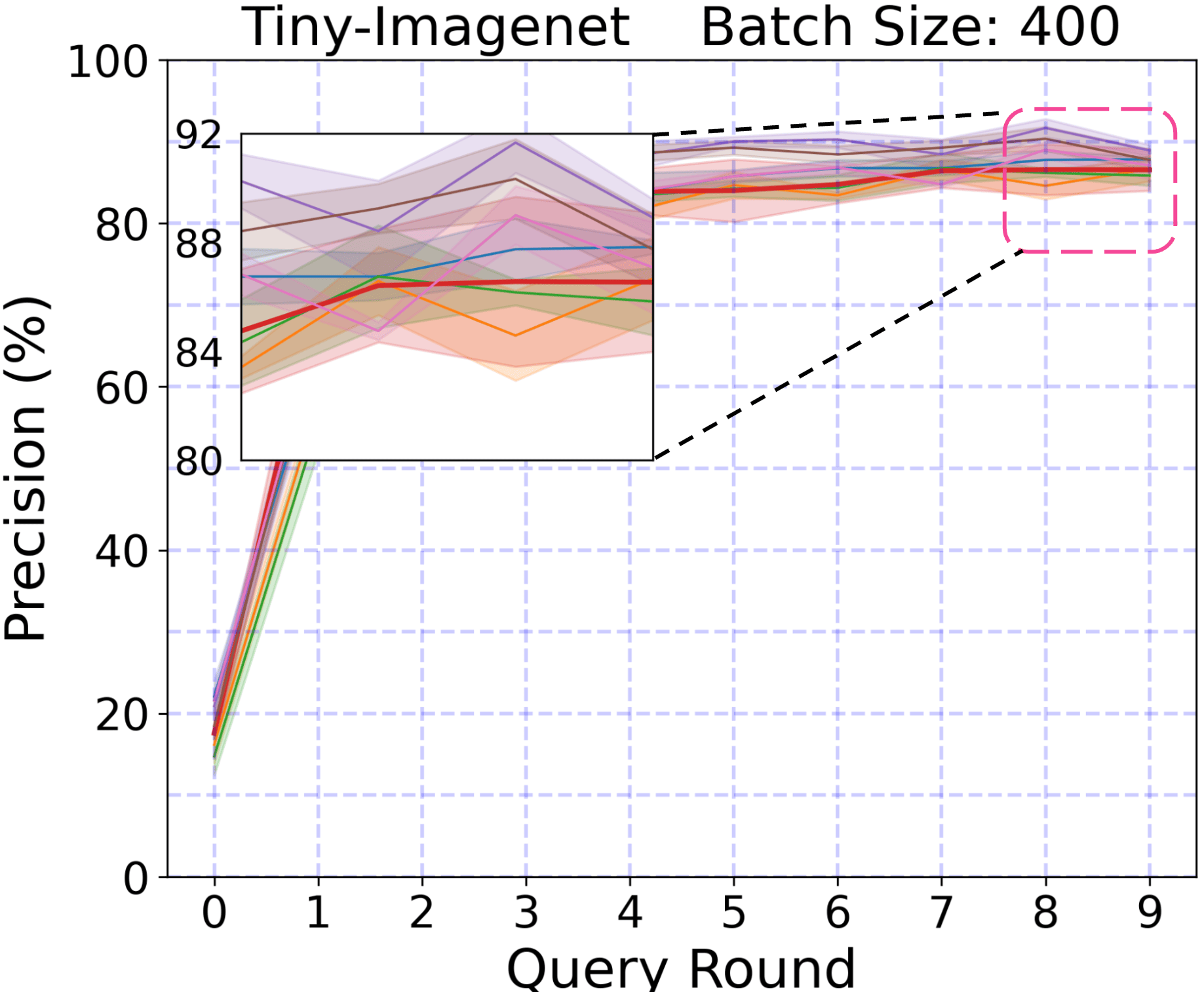} &
        \includegraphics[width=0.26\textwidth]{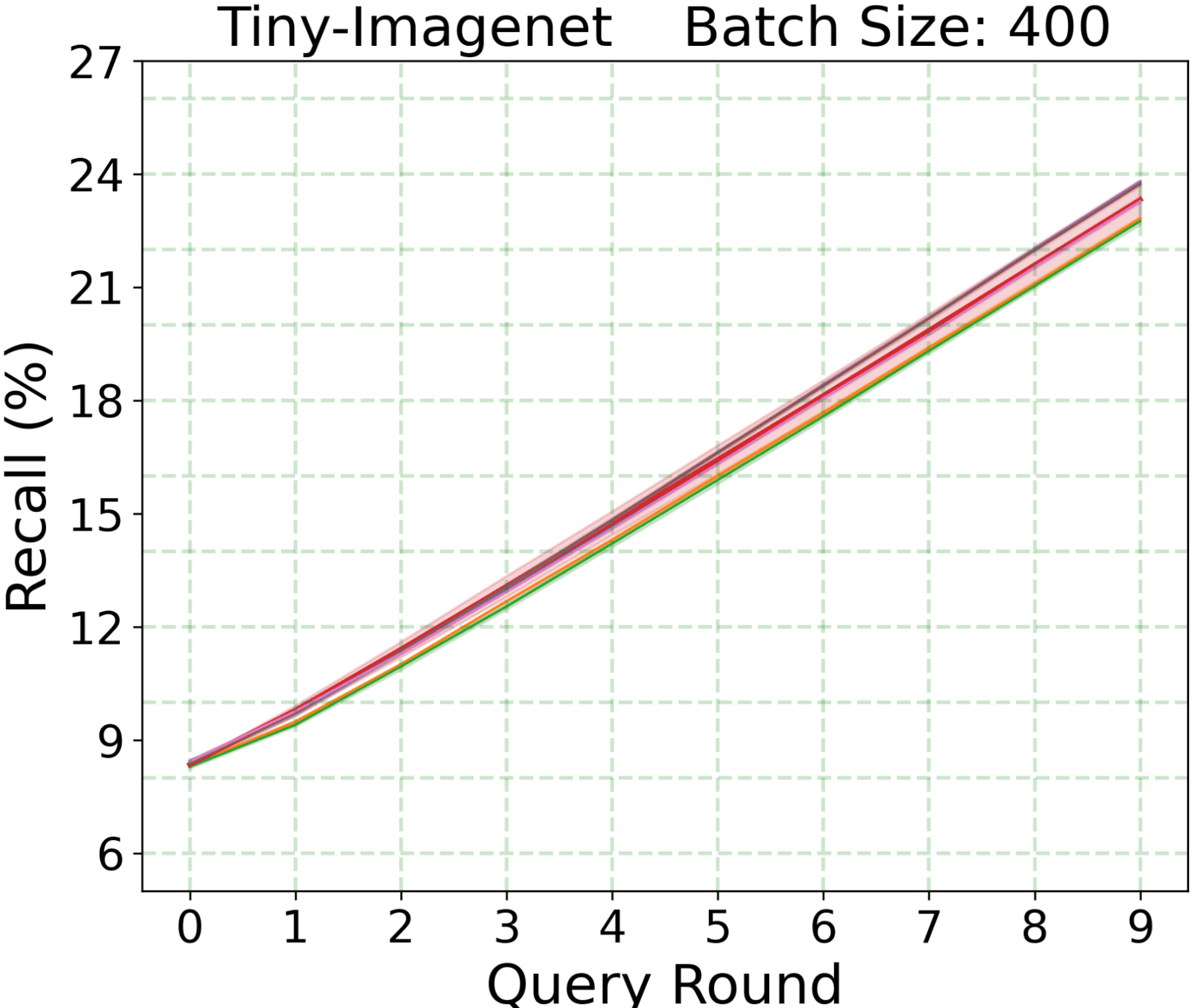} \\   
    \end{tabular}
    \begin{subfigure}{\textwidth}
        \centering
        \includegraphics[width=0.7\textwidth]{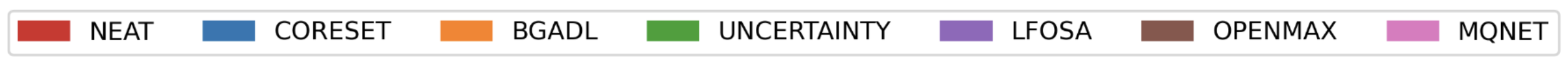}
    \end{subfigure}
    \caption{\textsc{Neat} is effective compared with other active learning methods for deep neural networks. }
    \label{fig:active}
\end{figure*}

The utilization of a pre-trained model inevitably introduces prior knowledge. Although this pre-trained model has not been directly trained on the target dataset, it is essential to examine its potential impact on other active learning methods. To this end, we conducted experiments involving the pre-trained model module across various baseline methods. In particular, we adopt a hybrid approach that combines \textit{data-centric known class detection} to initially identify potential known classes, followed by the application of different active learning methods using the data from these detected known classes. The empirical results indeed show the influence of the pre-trained model on classification accuracy. However, our proposed \textsc{Neat} method still surpasses other active learning techniques in terms of accuracy on the \textbf{Tiny-ImageNet}, as shown in Figure \ref{fig:active}.

\begin{table}[!h]
\centering
\small
\renewcommand{\arraystretch}{0.9} 
\begin{tabular}{lccc}
\toprule
Dataset & Model & Accuracy (avg ± std) \\
\midrule
\multirow{4}{*}{Tiny-Imagenet} & ResNet-50 & \(50.53 \pm 1.15\) \\
& ResNet-34 & \(50.17 \pm 0.20\) \\
& ResNet-18 & \(48.98 \pm 0.42\) \\
& CLIP & \(49.01 \pm 1.60\) \\
\midrule
\multirow{4}{*}{CIFAR100} & ResNet-50 & \(62.00 \pm 1.26\) \\
& ResNet-34 & \(62.73 \pm 0.45\) \\
& ResNet-18 & \(60.75 \pm 0.58\) \\
& CLIP & \(63.68 \pm 0.67\) \\
\midrule
\multirow{4}{*}{CIFAR10} & ResNet-50 & \(97.53 \pm 0.24\) \\
& ResNet-34 & \(97.48 \pm 0.19\) \\
& ResNet-18 & \(97.48 \pm 0.26\) \\
& CLIP & \(98.15 \pm 0.14\) \\
\bottomrule
\end{tabular}

\caption{\textsc{Neat} is robust to the choices of pre-trained models.}
\label{table:pre-trained}
\end{table}

\begin{table}[!h]
\small
\centering
\renewcommand{\arraystretch}{0.9} 
\begin{tabular}{lccc}
\toprule
Dataset & \# of Neighbors & Accuracy (avg ± std) \\
\midrule
\multirow{4}{*}{Tiny-Imagenet} & 5 & \(47.23 \pm 1.21\) \\
& 10 & \(49.01 \pm 1.60\) \\
& 15 & \(49.02 \pm 0.61\) \\
& 20 & \(50.73 \pm 0.29\) \\
\midrule
\multirow{4}{*}{CIFAR100} & 5 & \(62.63 \pm 0.66\) \\
& 10 & \(63.68 \pm 0.67\) \\
& 15 & \(62.77 \pm 0.84\) \\
& 20 & \(60.85 \pm 1.70\) \\
\midrule
\multirow{4}{*}{CIFAR10} & 5 & \(97.95 \pm 0.27\) \\
& 10 & \(98.15 \pm 0.14\) \\
& 15 & \(97.75 \pm 0.14\) \\
& 20 & \(97.85 \pm 0.21\) \\
\bottomrule
\end{tabular}
\caption{The impact of number of neighbors on the final performance of \textsc{Neat}.}

\label{table:neighbors} 
\end{table}

\subsubsection{Computational Efficiency}  Our approach is more computationally efficient compared to learning-based open-set active learning methods, such as \textsc{LfOSA}. This efficiency is attributed to the fact that learning-based methods like \textsc{LfOSA} require the additional training of a detection network to discern known and unknown classes. Consequently, the training time for our method on the \textbf{Tiny-ImageNet} is 88 minutes on an NVIDIA A5000 GPU, while the \textsc{LfOSA} method requires 156 minutes. For the \textbf{CIFAR100}, our method takes 21 minutes, compared to LFOSA's 59 minutes, and for the \textbf{CIFAR10}, our method requires 17 minutes, while LFOSA needs 25 minutes. Overall, we can observe that \textsc{Neat} is much faster than LFOSA.

\subsection{Ablation Studies}
\noindent \textbf{Impact of feature quality.} 
We investigate the impact of different pre-trained models for feature extraction. The quality of the features is a crucial factor that may affect the label clusterability of the dataset. 
Specifically, we consider pre-trained ResNet-18, ResNet-34, and ResNet-50 on ImageNet, in addition to CLIP. Figure \ref{fig:baseline} demonstrates that \textsc{Neat} achieves significantly better accuracy than the baseline active learning methods, regardless of the pre-trained models used. This highlights the robustness of \textsc{Neat} in detecting known classes. It's worth noting that while CLIP features achieve better accuracy on \textbf{CIFAR-10} and \textbf{CIFAR-100} than ResNets, the accuracy on \textbf{Tiny-ImageNet} is lower. This could be because the ResNets are pre-trained on ImageNet, and their features are better suited for Tiny-ImageNet. However, large language models like CLIP can generally provide high-quality features that are useful across different datasets.

\noindent \textbf{Influence of number of neighbors.} We further investigate how the number of neighbors impacts the detection of data-centric known classes. We consider different values of $K$ (5, 10, 15, 20) and present the results in Table \ref{table:neighbors}. We observe that although a smaller value of $K$ (e.g., $K=5$) leads to slightly worse performance, other choices of $K$ yield similar results. This suggests that a smaller $K$ only captures the local feature distribution of the target sample, which not provide a good characterization of the underlying feature space. Ablation studies of query batch size and the impact of different classification models are included in the Appendix.

\begin{figure}[!t]
\centering
\includegraphics[width=0.6\linewidth]{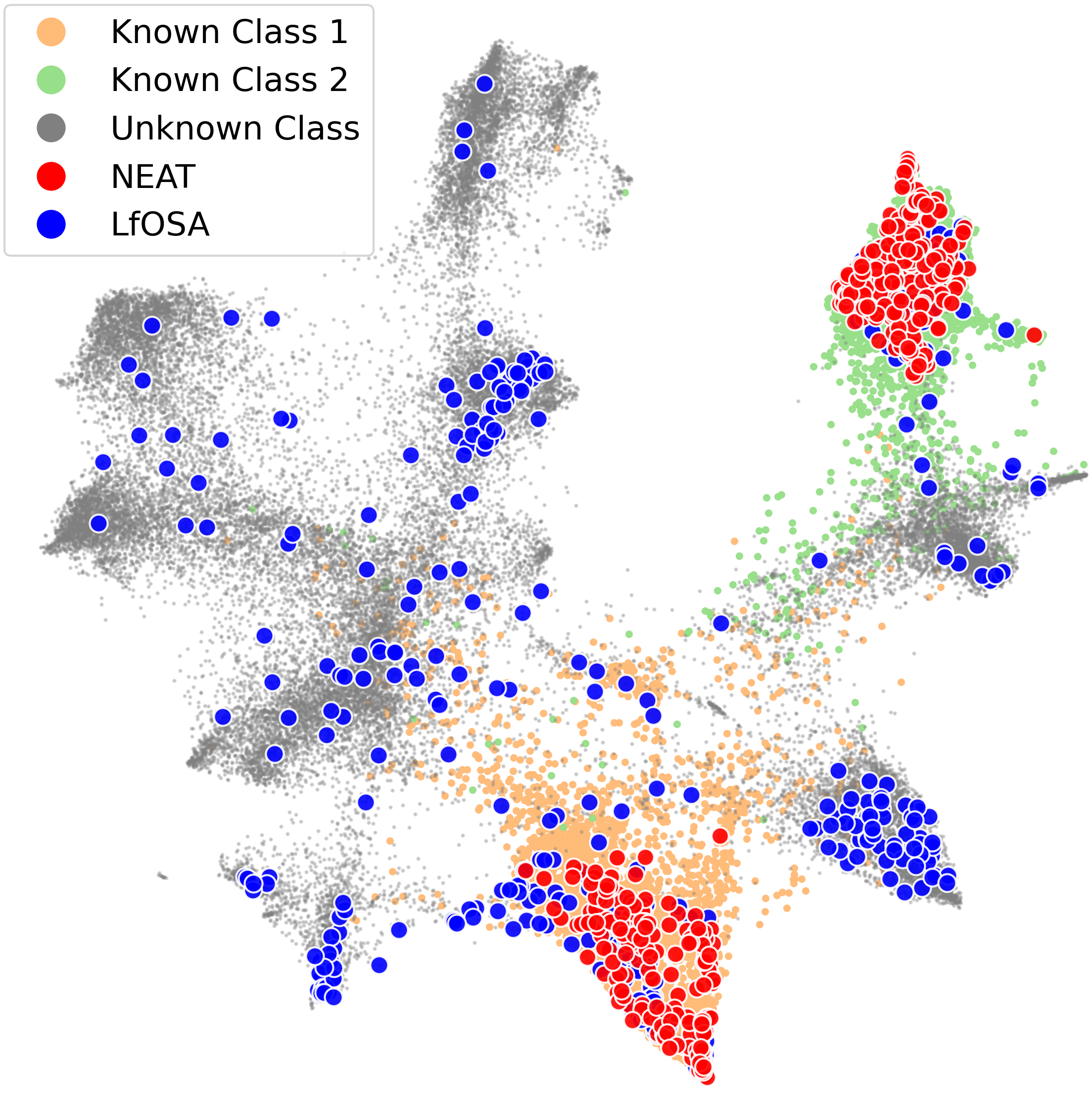}
\caption{\textsc{Neat} can accurately identify known classes from the unlabeled pool.}
    \label{fig:tsne}
\end{figure}


\noindent \textbf{Visualization.} We used t-SNE \cite{van2008visualizing} to visualize the features produced by CLIP on \textbf{CIFAR-10}, focusing on the samples selected by \textsc{Neat} and \textsc{LfOSA} \cite{ning2022active}. We randomly select a query round for plotting. Our results show \textsc{LfOSA} selects many samples from unknown classes, explaining its low precision. Alternatively, nearly all samples \textsc{Neat} selects are from known classes, demonstrating its effectiveness in the active open-set annotation problem.

\section{Conclusion}
In this paper, we propose a solution to the practical challenge of maintaining high recall when identifying examples of known classes for target model training from a massive unlabeled open-set. To address this challenge, we introduce a data-centric active learning method called \textsc{Neat}, which utilizes existing large language models to identify known classes in the unlabeled pool. Our proposed method offers several advantages over traditional active learning methods. Specifically, \textsc{Neat} uses the CLIP model to extract features, which eliminates the need to train a separate detector. Additionally, our approach achieves improved results with low query numbers, resulting in cost savings on labeling. 

\section{Acknowledgments}
This work was supported by a startup funding by the University of Texas at Dallas.

\bibliography{aaai24}
\clearpage
\section{Theoretical analysis}
We give an upper bound on the detection error of the proposed known class detection. The first step is to understand in what condition our method will make a mistake. Given the sample $\mathbf{x}$, although all of its $K$-nearest neighbors $\{ N_1(\mathbf{x}), N_2(\mathbf{x}), ..., N_K(\mathbf{x})\}$ belong to known classes, it is still possible that the sample is from the unknown classes. This can caused by the randomness in the labeling process and the quality of the features.

To capture the randomness caused by the quality of the features, we introduce the following assumption,

\begin{assumption}
There exists a constant $C > 0$ and $0 < \alpha < 1$ such that for any $\mathbf{x}$ and  $\mathbf{x}'$,
\begin{equation} 
    \mathbb{P}( y_{true}(\mathbf{x}) \neq y_{true}(\mathbf{x}')) \le C \rho(\mathbf{x}, \mathbf{x}')^\alpha
    \label{assump: dist1}
\end{equation}
\end{assumption}

\noindent Here, $\rho(\mathbf{x}, \mathbf{x}')$ represents the distance between samples $\mathbf{x}$ and $\mathbf{x}'$, while $y_{true}(\mathbf{x})$ and $y_{true}(\mathbf{x}')$ correspond to the true labels of $\mathbf{x}$ and $\mathbf{x}'$, respectively. We use $r_K(\mathbf{x})$ to denote the radius of the ball centered at $\mathbf{x}$ such that $\forall$ $\mathbf{x}'$ in the $K$-nearest neighbors of $\mathbf{x}$, $\rho(\mathbf{x}, \mathbf{x}') \le r_K(\mathbf{x})$. 

To capture the randomness caused by the labeling process, we introduce the following assumption,

\begin{assumption}
For any $\mathbf{x}$, the labeling error is upper-bounded by a small constant $e$,
\begin{equation} 
    \mathbb{P}( y(\mathbf{x}) \neq y_{true}(\mathbf{x})) \le e
    \label{assump: dist2}
\end{equation}

\end{assumption}

Then, we can establish an upper bound for the detection error as follows,
\begin{theorem} Given the assumption 0.1. and 0.2. and the number of neighbors $K$, the probability of making a detection error is upper-bounded as,
\begin{align}
\begin{split}
\mathbb{P}(\textnormal{Error} | & K) \le  \sum_{k=\ceil{\frac{K - 1}{2}} + 1}^K \binom Kk e^k(1-e)^{K-k}  C^k r_K(\mathbf{x})^{\alpha k} \\
& + \sum_{k=0}^{  \floor{\frac{K+1}{2}}-1 } \binom Kk e^k(1-e)^{K-k}  C^{(K-k)} r_K(\mathbf{x})^{\alpha (K-k)}
\end{split}
\end{align}
\end{theorem}

\noindent Proof: There are two scenarios in which the detection method may produce errors: 1) When a majority of the neighboring instances are assigned incorrect labels and 2) when a majority of the neighboring instances possess accurate labels. However, due to the feature quality, the possibility remains that the label of the example $\mathbf{x}$ originates from unknown classes.
In the first scenario, we account for the likelihood of encountering a minimum of $\lceil\frac{K - 1}{2}\rceil + 1$ neighbors with wrong labels. Additionally, even if a significant number of neighbors belong to unknown classes, the chance persists that the sample originates from known classes. In the second scenario, we consider the possibility of having a maximum of $\lfloor\frac{K+1}{2}\rfloor-1$ neighbors with incorrect labels. Similarly, even if most neighbors are associated with known classes, the potential remains that the sample comes from unknown classes.

\section{Detailed Active Open-Set Annotation Setup}

\subsection{Datasets}
In our experiment, we use the \textbf{CIFAR10}, \textbf{CIFAR100}, and \textbf{Tiny-ImageNet} to test our method. The \textbf{CIFAR10} consists of 60,000 images in total, each being a 32x32 pixels images. This dataset is divided into 10 classes, with 6,000 images per class. For the purpose of our experiments, we randomly select 2 out of these 10 classes as known classes, which include a variety of categories such as airplanes, cars, and cats. Out of the total 60,000 images, 50,000 are used as a training dataset while the remaining 10,000 serve as a test dataset. To initialize our experiment, we randomly select 1 percent of the known classes' training dataset's images, resulting in 100 images used for the classifier model's training.

The \textbf{CIFAR100}, similar to the \textbf{CIFAR10}, consists of 60,000 32x32 pixels images. However, it has a different configuration which has 100 classes, each containing 600 images which has 500 for training and 100 for testing. As same as the \textbf{CIFAR10}, 50,000 images in the CIFAR100 dataset are allocated for training purposes and the remaining 10,000 are used for testing. Furthermore, the \textbf{CIFAR100} introduces an additional layer of categorization; the 100 classes are grouped into 20 superclasses, known as 'coarse labels'. Each coarse label corresponds to 5 classes. For our experiment, we randomly select 8\% of the images from the known class as our starting point, which equates to 80 images.

The \textbf{Tiny-ImageNet} comprises 20 classes, with each class containing 500 training images, 50 validation images, and 50 test images. Each image is 64x64 pixels, which in total has 100,000 images making this dataset significantly larger than either \textbf{CIFAR10} or \textbf{CIFAR100} in terms of image number and resolution. Tiny-ImageNet is a subset of the ILSVRC-2012 classification dataset, representing a scaled-down version of the larger ImageNet dataset, which has over 14 million images. For our experiment, we randomly select 8\% of the images from the known class to initialize our test, which also equals to 80 images. Due to the absence of labels in the \textbf{Tiny-ImageNet} test set, we utilize the validation set as our test set.

\subsection{Active Open-Set Annotation}

After the completion of the initialization stage, the subsequent stage is the query stage. During this stage, the images belonging to the remaining unselected known classes after the initialization, along with those from unknown classes, are combined to form a new, unlabeled dataset. The next process is different for learning-based methods and non-learning-based methods. For the learning-based active learning method, which has an additional detector network, the images randomly selected at the initialization stage will be used to train. Then the newly trained detector network will be used to predict the images from the unlabeled dataset because the features or activation values will be used for the known and unknown class selection. For the non-learning-based method, the unlabeled images will go through their own individual selection process. 

At the query stage, both active learning methods are required to select \textsc{K} number of images, which each method predicted to have the highest possibility from known classes. Next is the auto-annotation process, where the selected images will be categorized into known classes and unknown classes based on their true label. Besides the index of selected images, the query step will also output the selection precision and recall. 

The final process of the query stage is also divided into different pipelines regarding whether the current active learning method is learning-based or not. For the learning-based active learning method, the selected images that are truly from known classes will be added to the previous labeled dataset, and the images from the unknown classes will also be added to the labeled dataset, except their label will be changed to the number of known classes plus one to indicate an unknown class. For the non-learning-based method, only images from known classes will be added to the labeled dataset. The above querying process will repeat a \textsc{Q} number of times, and the last query's classifier's performance will be used to evaluate the current active learning method's effectiveness.

\section{Further Ablation Studies}

\subsection{Query batch size}
\begin{figure*}[p]
    \centering
    \setlength{\tabcolsep}{2pt}
    \begin{tabular}{ccc}
        \includegraphics[width=0.25\textwidth]{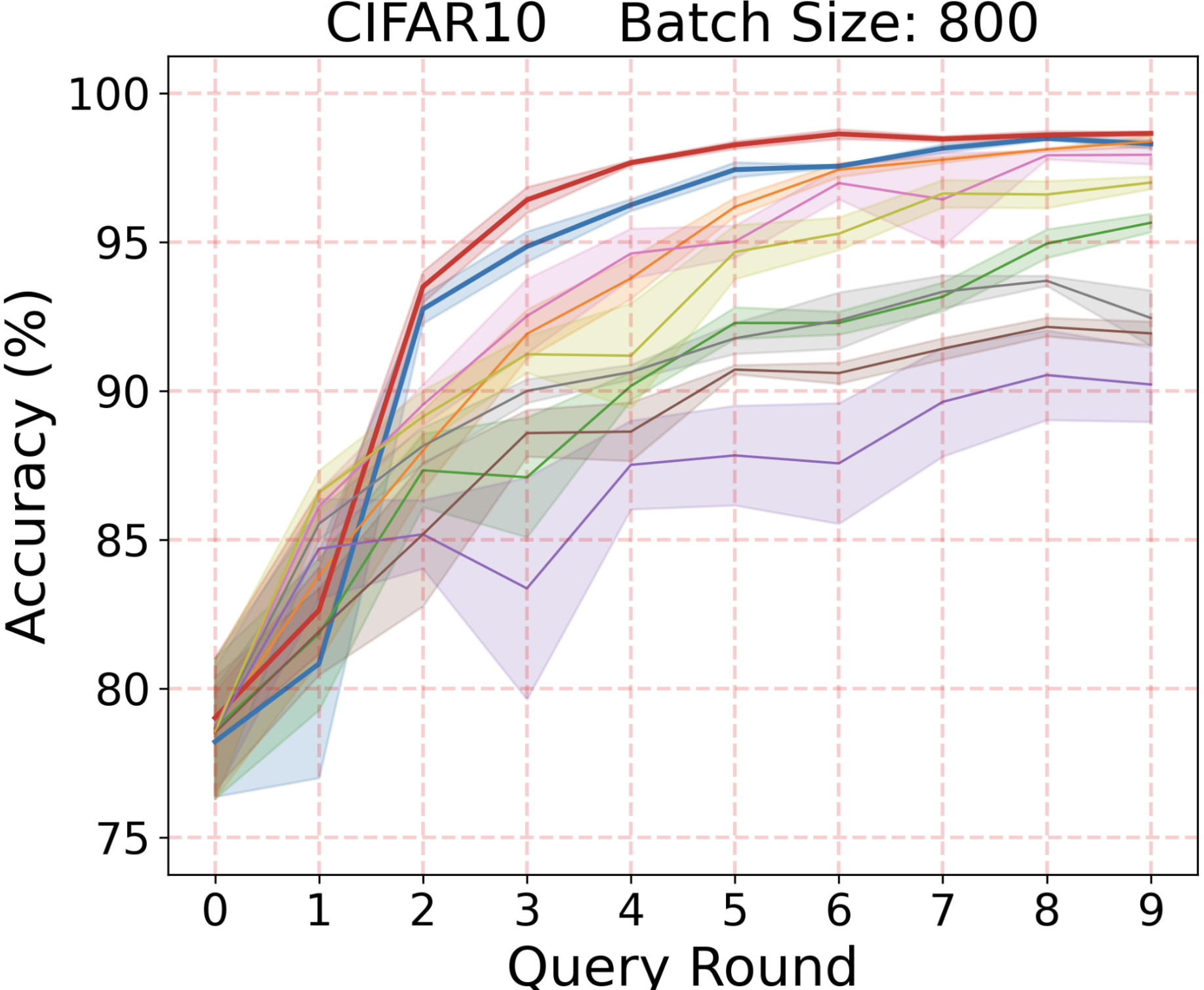} &
        \includegraphics[width=0.25\textwidth]{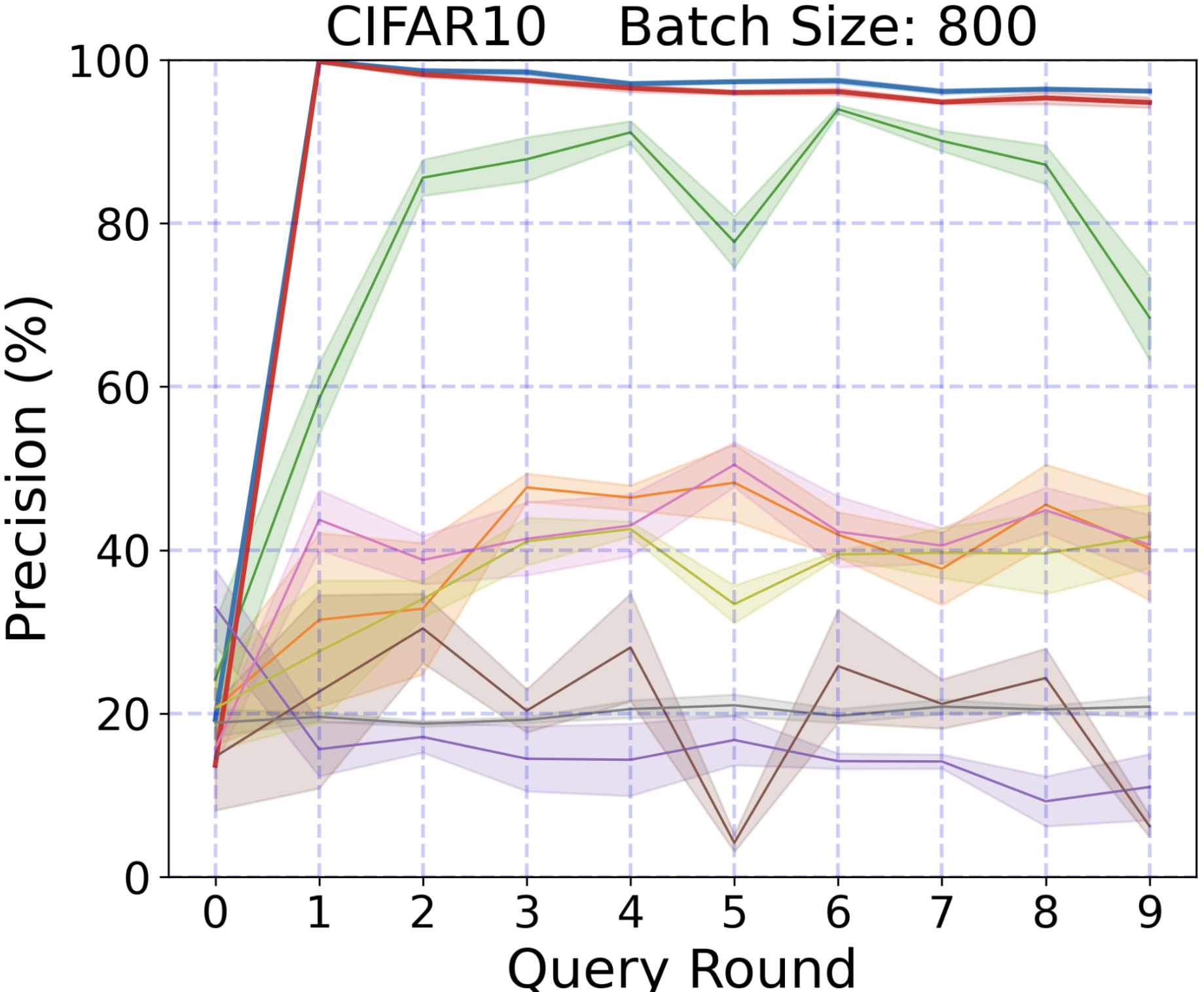} &
        \includegraphics[width=0.25\textwidth]{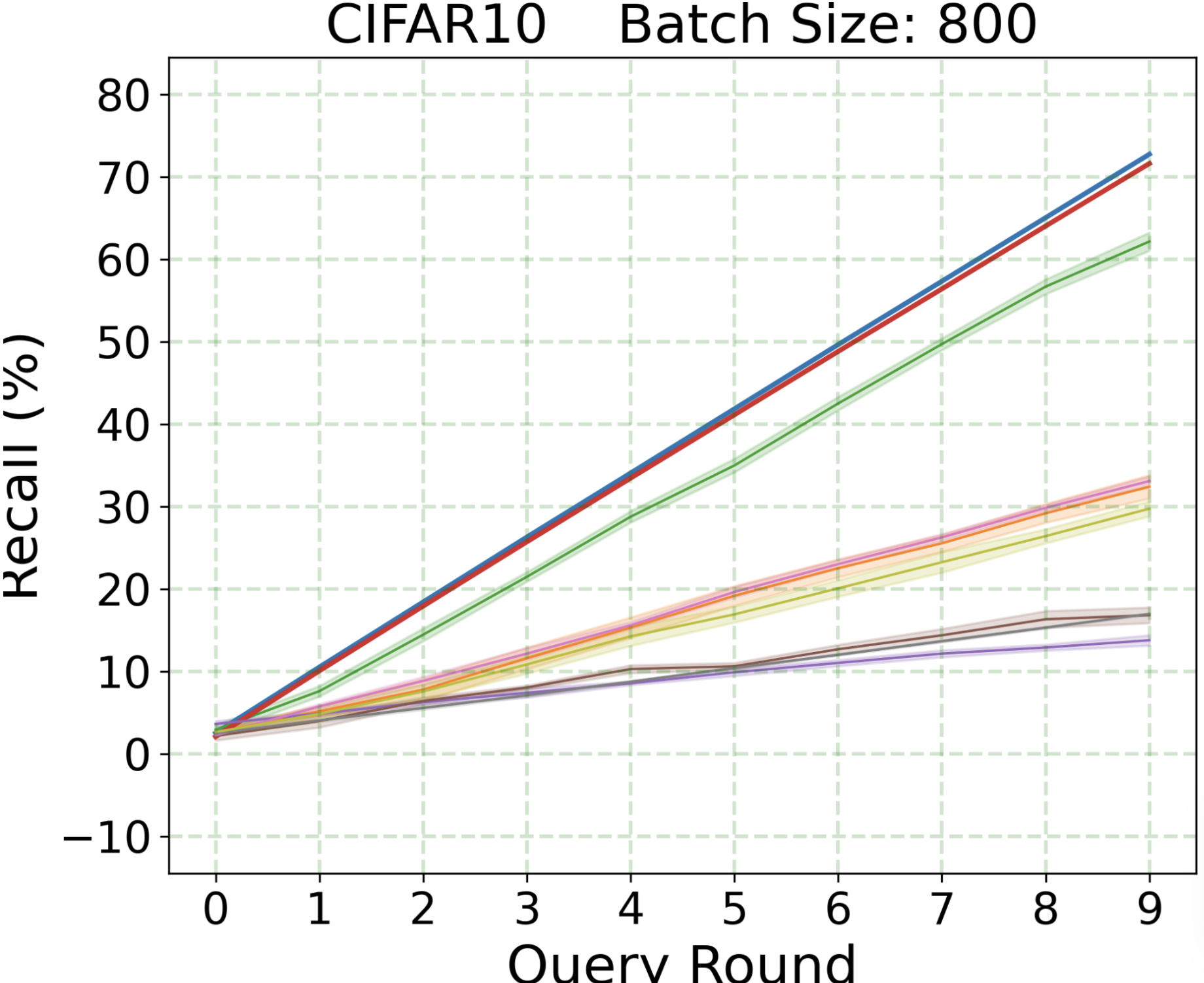} \\
        \includegraphics[width=0.25\textwidth]{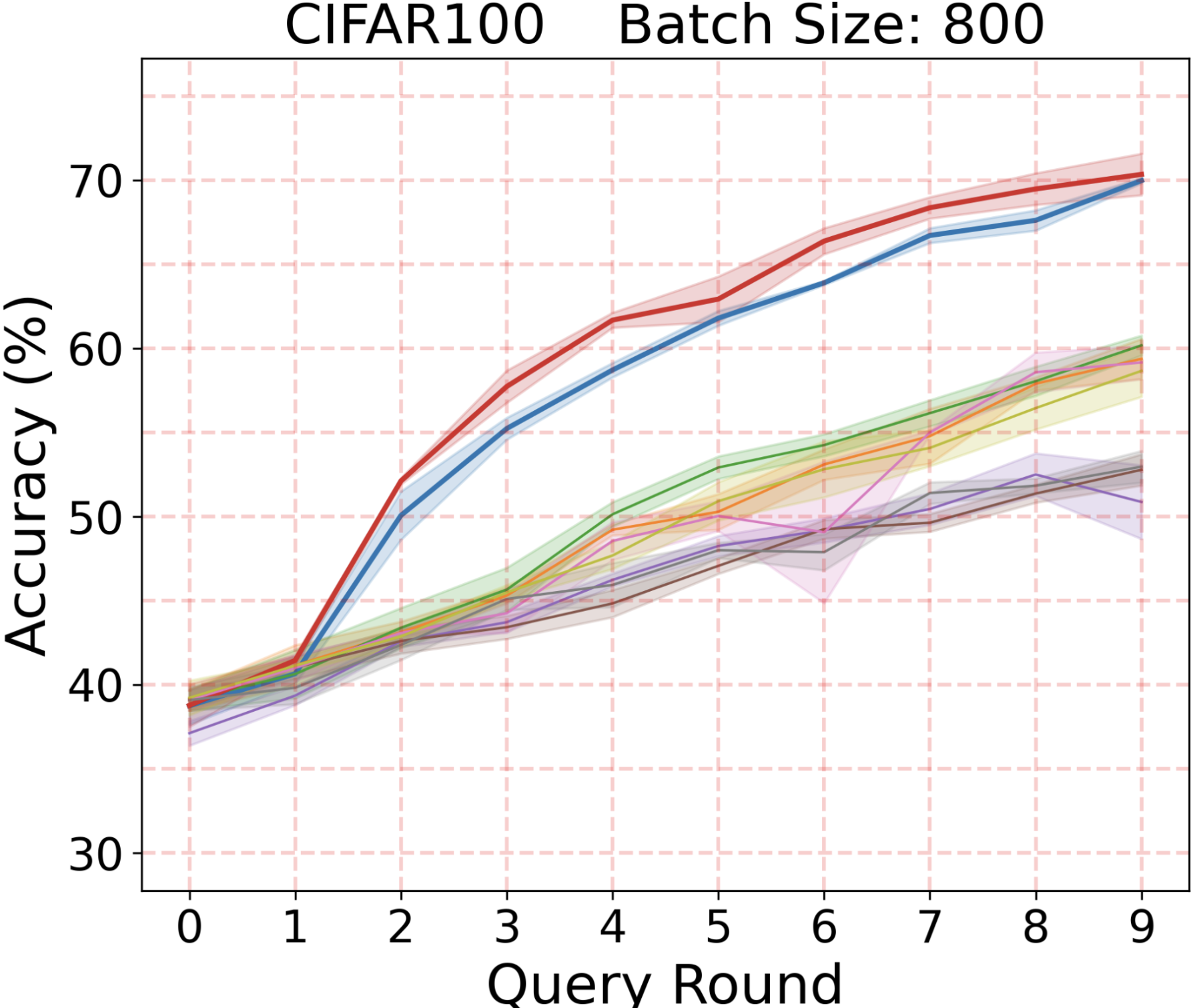} &
        \includegraphics[width=0.25\textwidth]{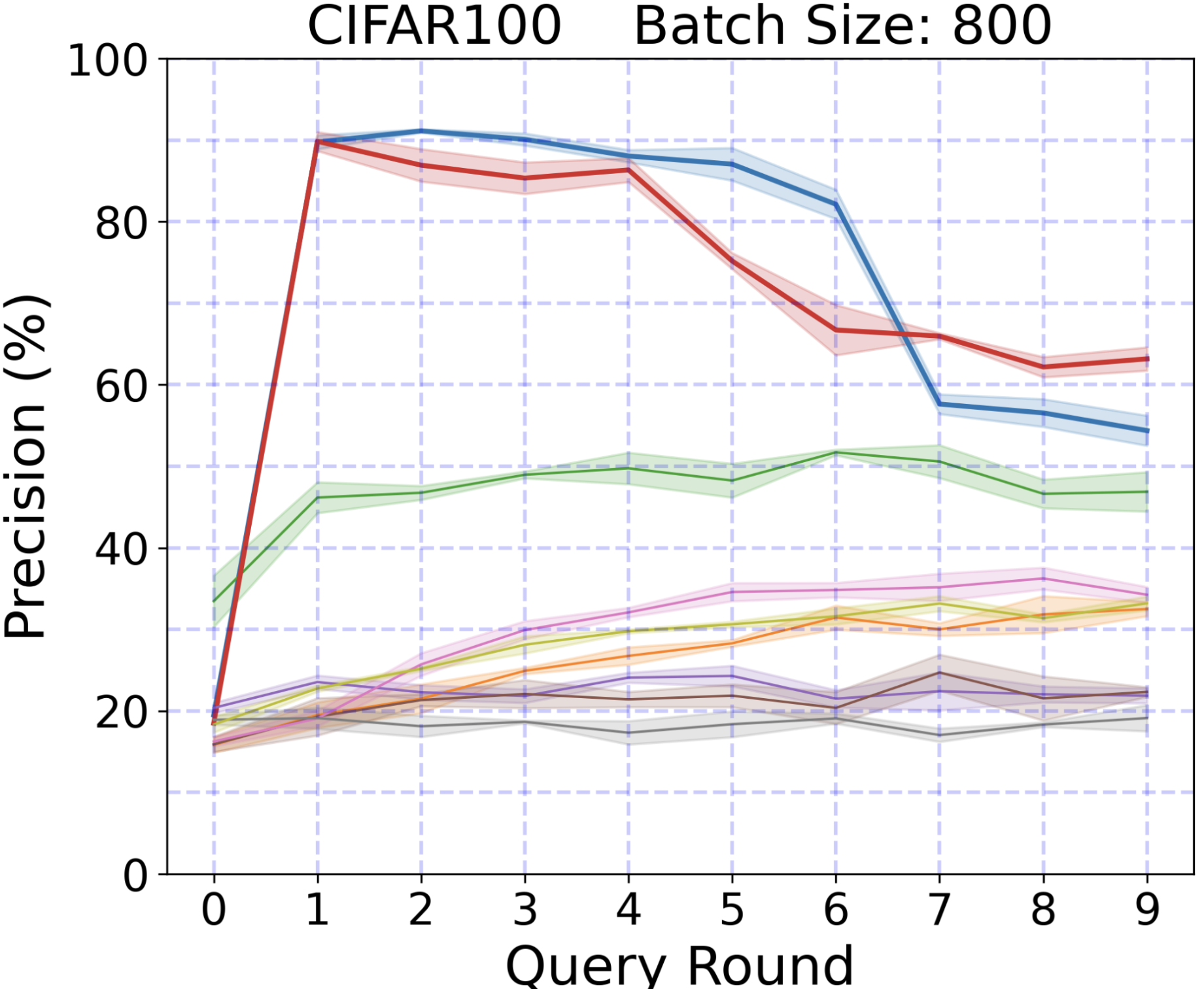} &
        \includegraphics[width=0.25\textwidth]{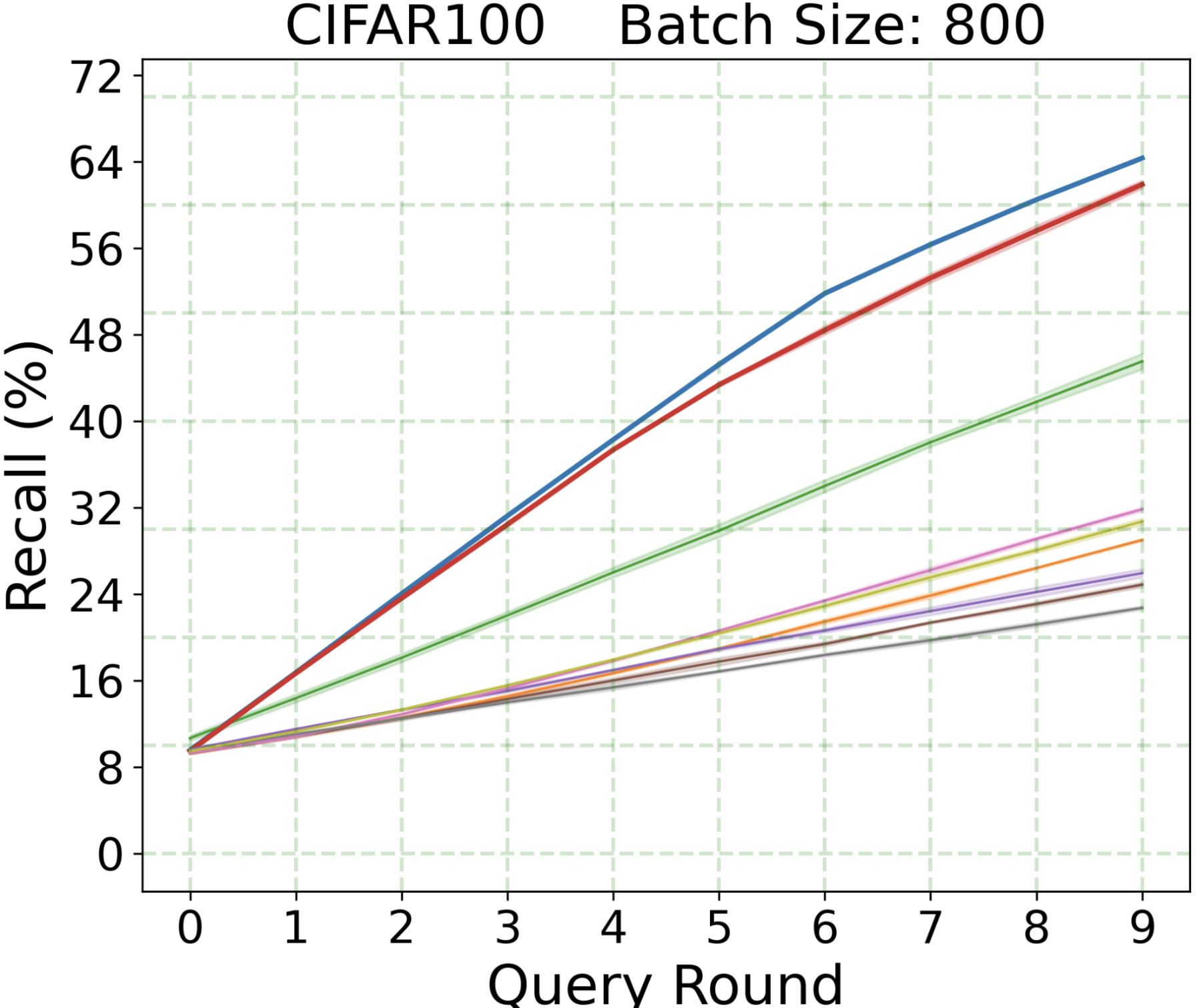} \\    
        \includegraphics[width=0.25\textwidth]{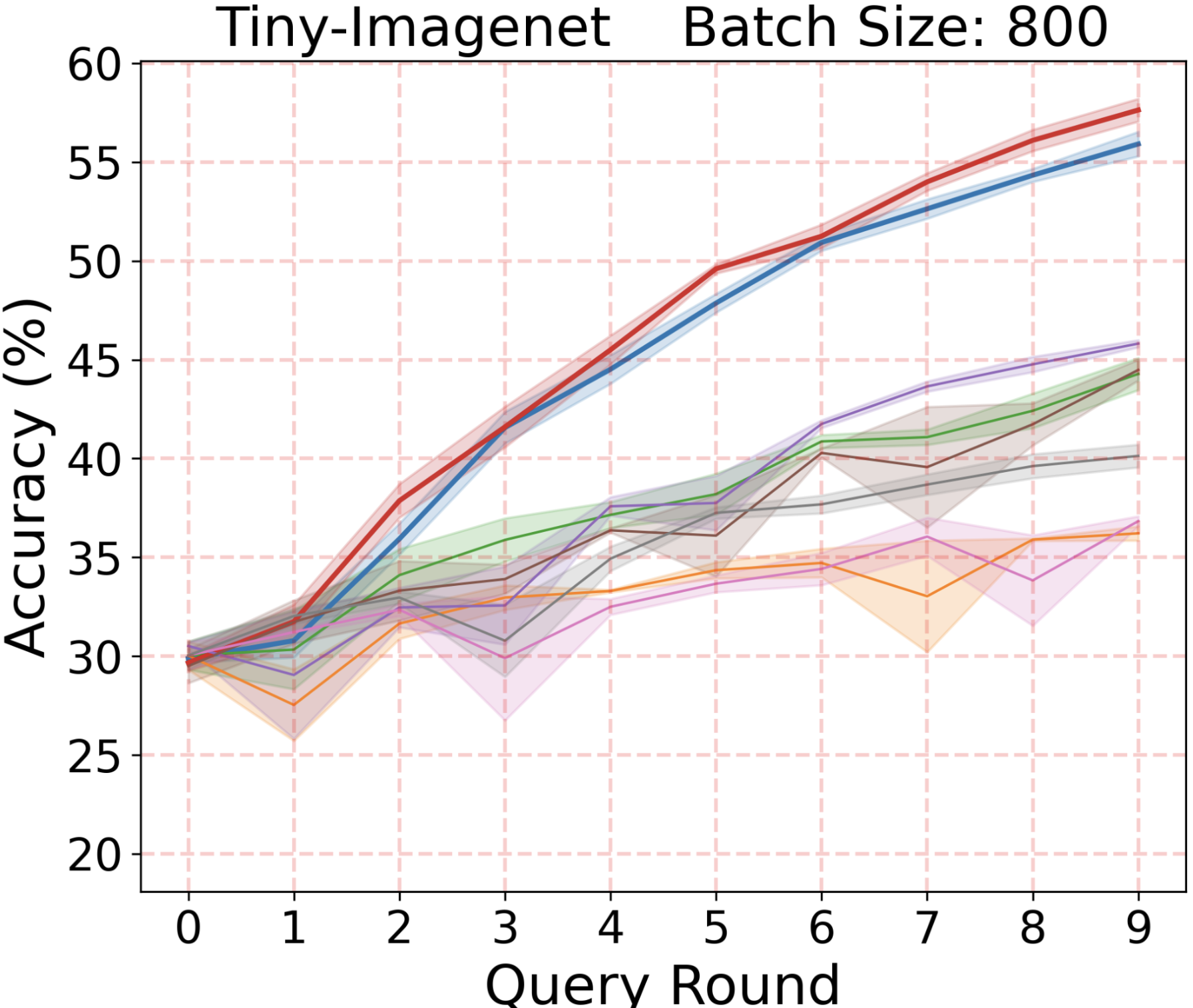} &
        \includegraphics[width=0.25\textwidth]{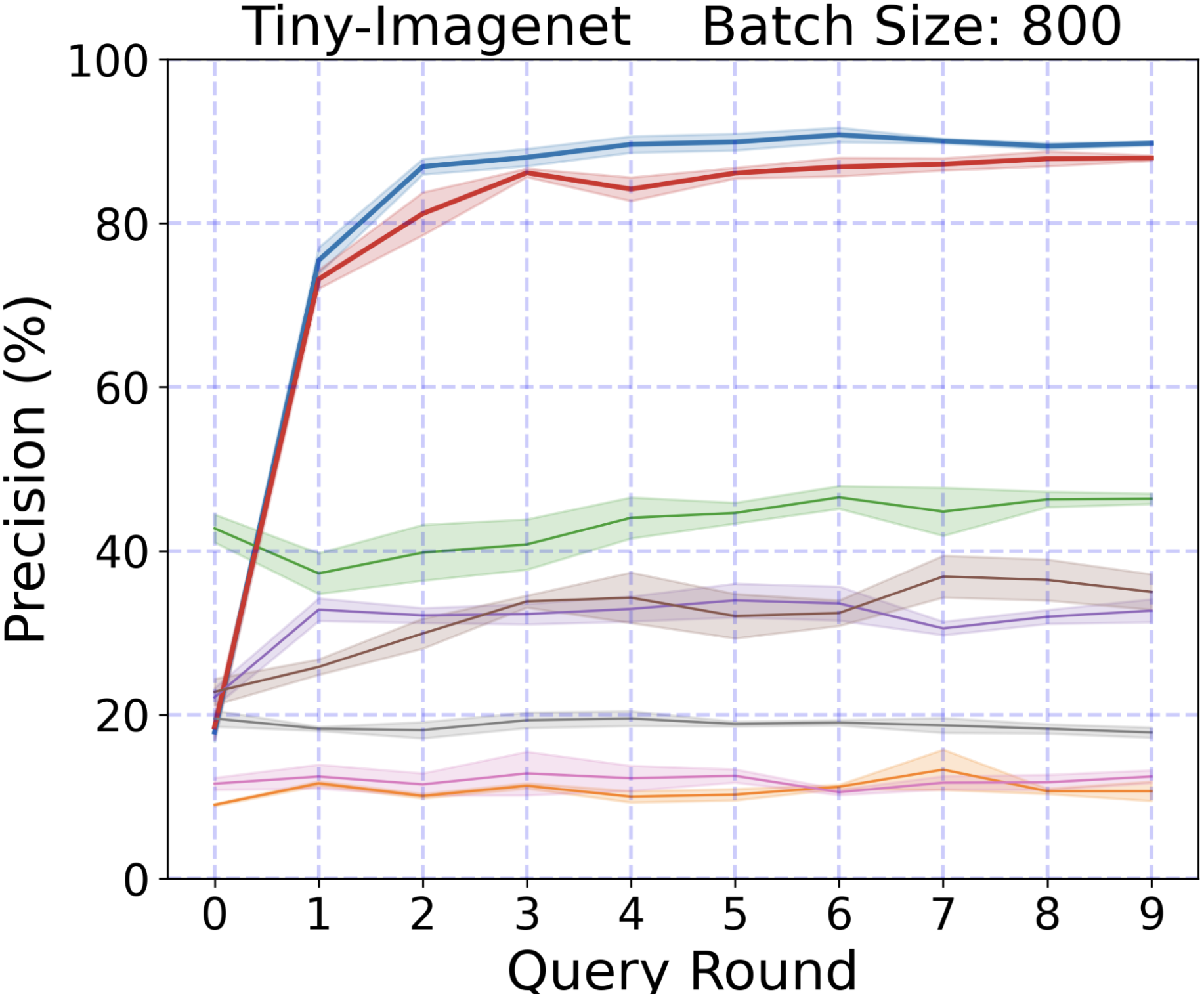} &
        \includegraphics[width=0.25\textwidth]{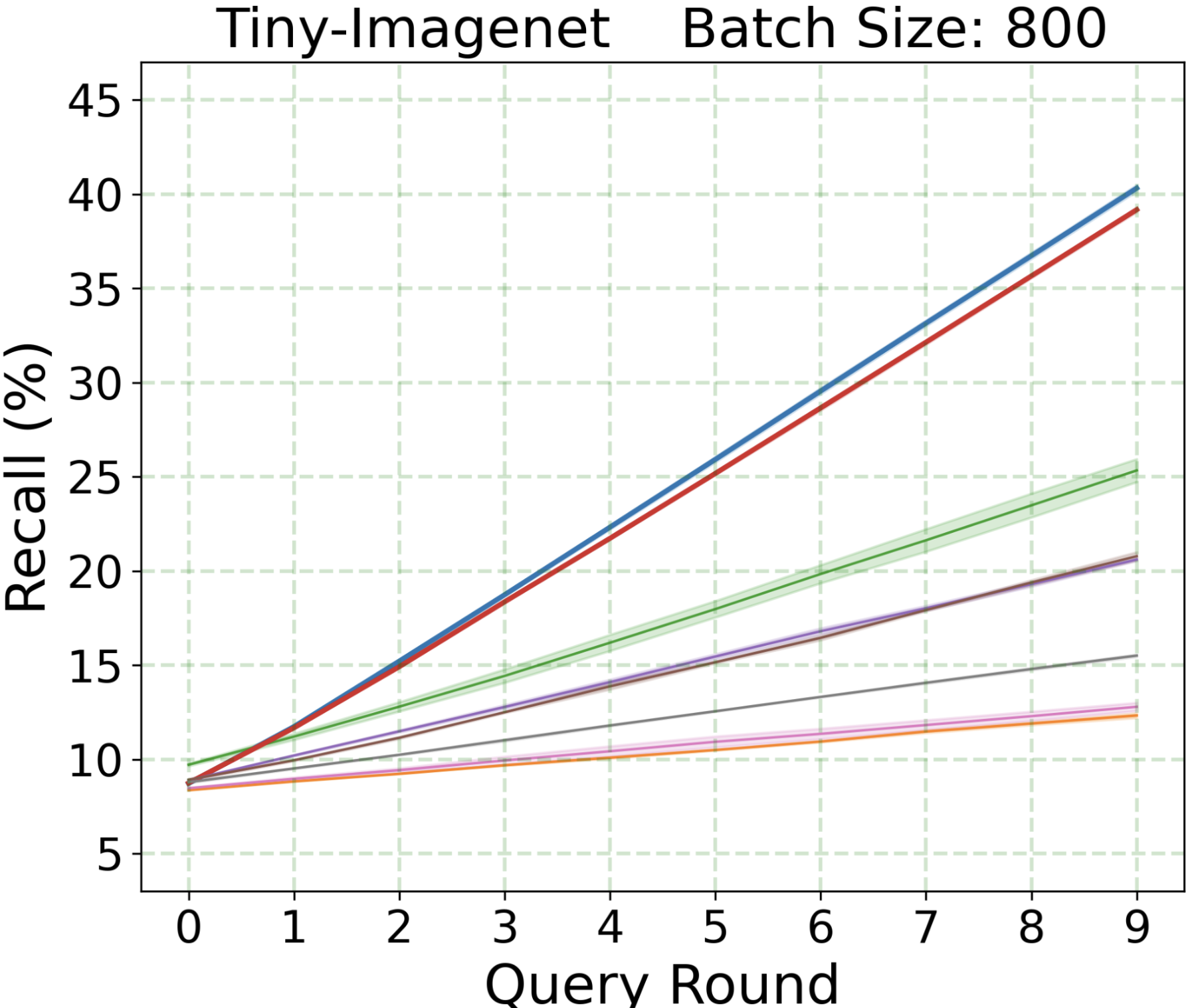} \\
        \multicolumn{3}{c}{\includegraphics[width=0.7\textwidth]{baseline/legund.png}} \\              
    \end{tabular}
    \caption{\textsc{Neat} achieves higher precision, recall and accuracy compared with existing active learning methods for active open-set annotation. We evaluate \textsc{Neat} and the baseline active learning methods on \textbf{CIFAR10}, \textbf{CIFAR100} and \textbf{Tiny-ImageNet} based on accuracy, precision and recall. }
    \label{fig:baseline2} 
\end{figure*}

In addition to the experiments detailed in our main paper, where we used a query batch size of 400 images, we also present our experimental results for query batch sizes of 600 and 800. Our observations reveal that as the batch size increases, the accuracy of all active learning methods improves. However, our proposed method, \textsc{Neat}, outperforms other methods in terms of accuracy, precision, and recall across all cases. Particularly at smaller query batch sizes, our method achieves significantly higher accuracy compared to alternative approaches, especially at a batch size of 400. Furthermore, utilizing a batch size of 400 images reduces annotation costs by almost half compared to a batch size of 800. The reduction in batch size is crucial in open-set active learning because it reduces the annotation cost and computation cost at the same time. The outcomes can be observed in Figures \ref{fig:baseline} and \ref{fig:baseline2}.

\subsection{Network architectures}
\begin{figure*}[t]
    \centering
    \setlength{\tabcolsep}{2pt}
    
        \begin{tabular}{ccc}
            \includegraphics[width=0.28\textwidth]{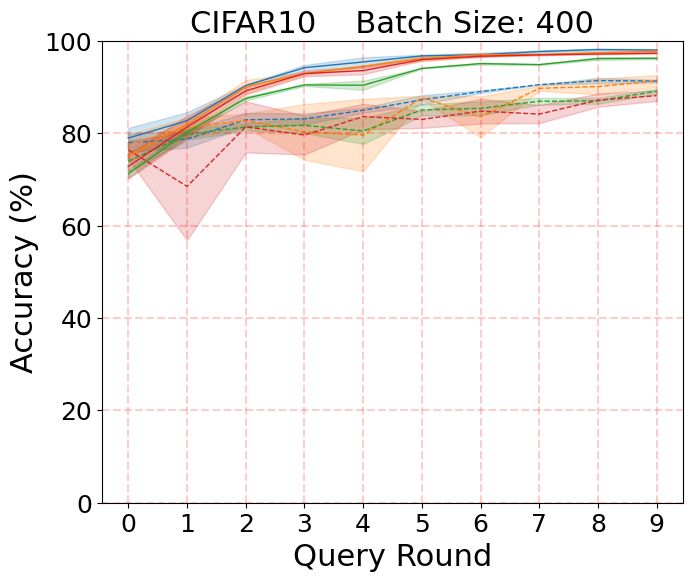} &
            \includegraphics[width=0.28\textwidth]{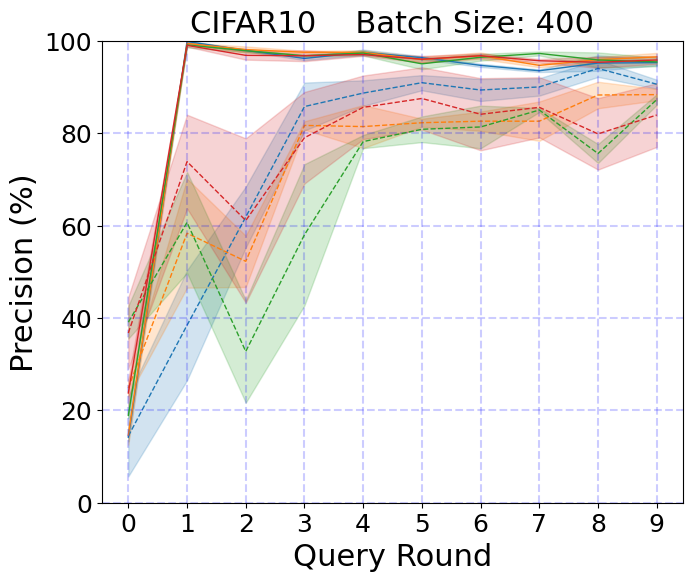} &
            \includegraphics[width=0.28\textwidth]{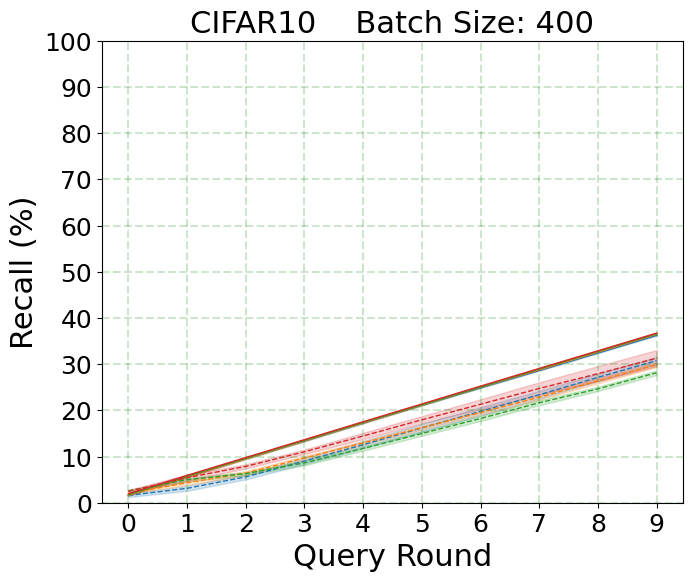} \\
            \includegraphics[width=0.28\textwidth]{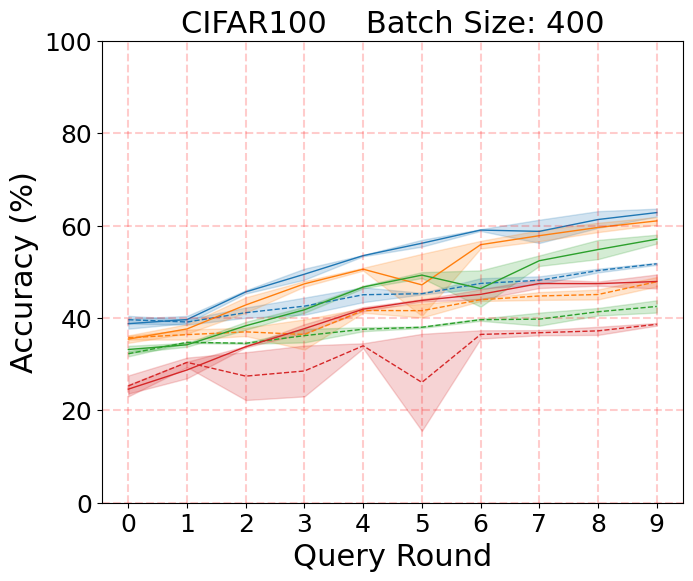} &
            \includegraphics[width=0.28\textwidth]{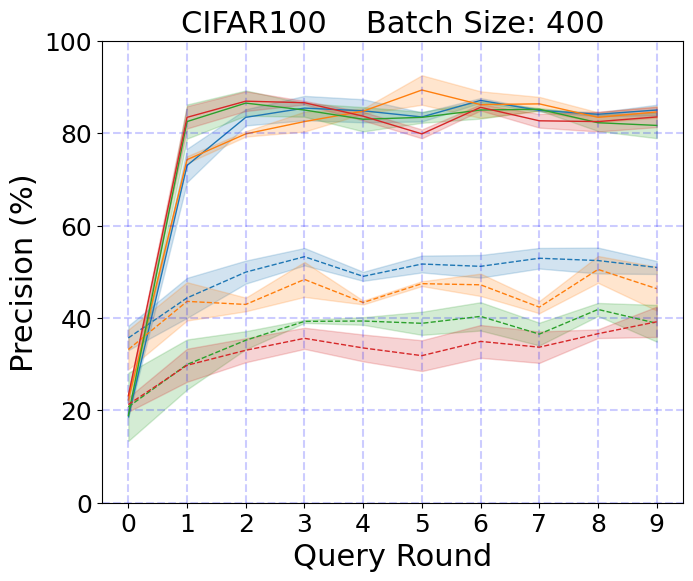} &
            \includegraphics[width=0.28\textwidth]{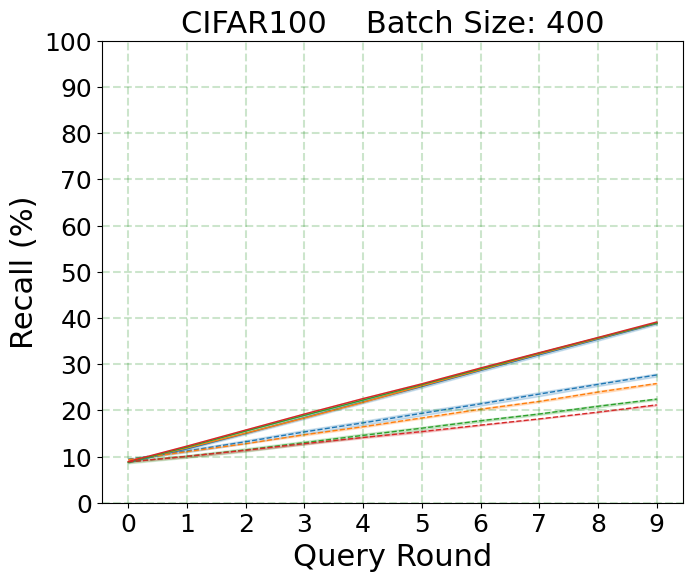} \\ 
            \multicolumn{3}{c}{
                \includegraphics[width=0.4\textwidth]{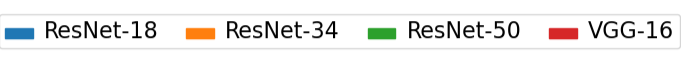}
                \includegraphics[width=0.4\textwidth]{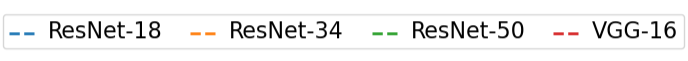}
            } \\   
        \end{tabular}
        \caption{\textsc{Neat} achieves higher precision, recall, and accuracy compared with LFOSA. We evaluate \textsc{Neat} and the LFOSA on \textbf{CIFAR10}, \textbf{CIFAR100} based on accuracy, precision, and recall. Specifically the \textbf{solid line} represents the result of \textsc{Neat} and the \textbf{dashed line} represents the result of  \textsc{LfOSA}.}
        \label{fig:structure}
    
\end{figure*}

\begin{figure}[t]
\centering
\begin{tabular}{cc}
   \includegraphics[width=1\linewidth]{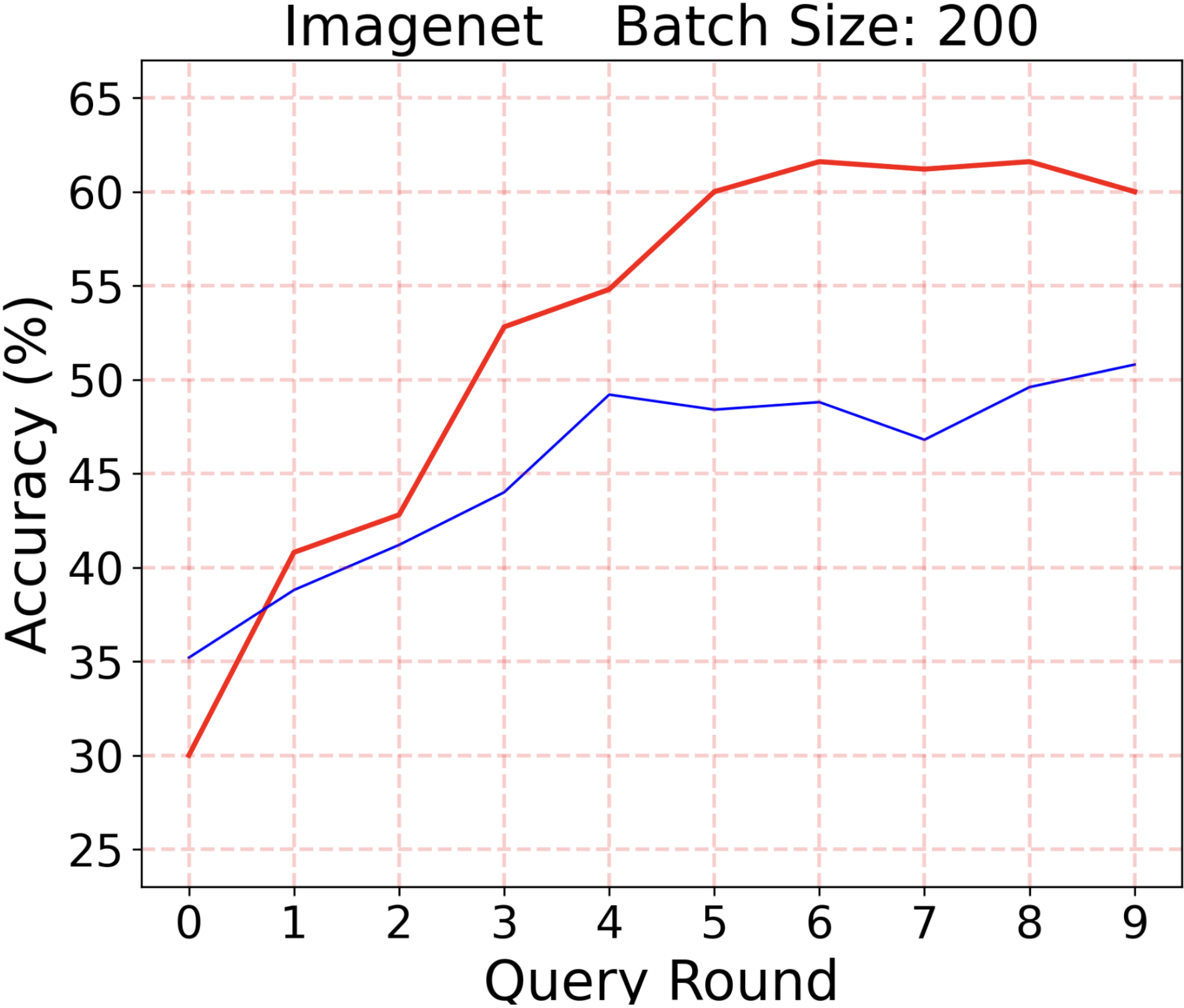}  &   \includegraphics[width=1.01\linewidth]{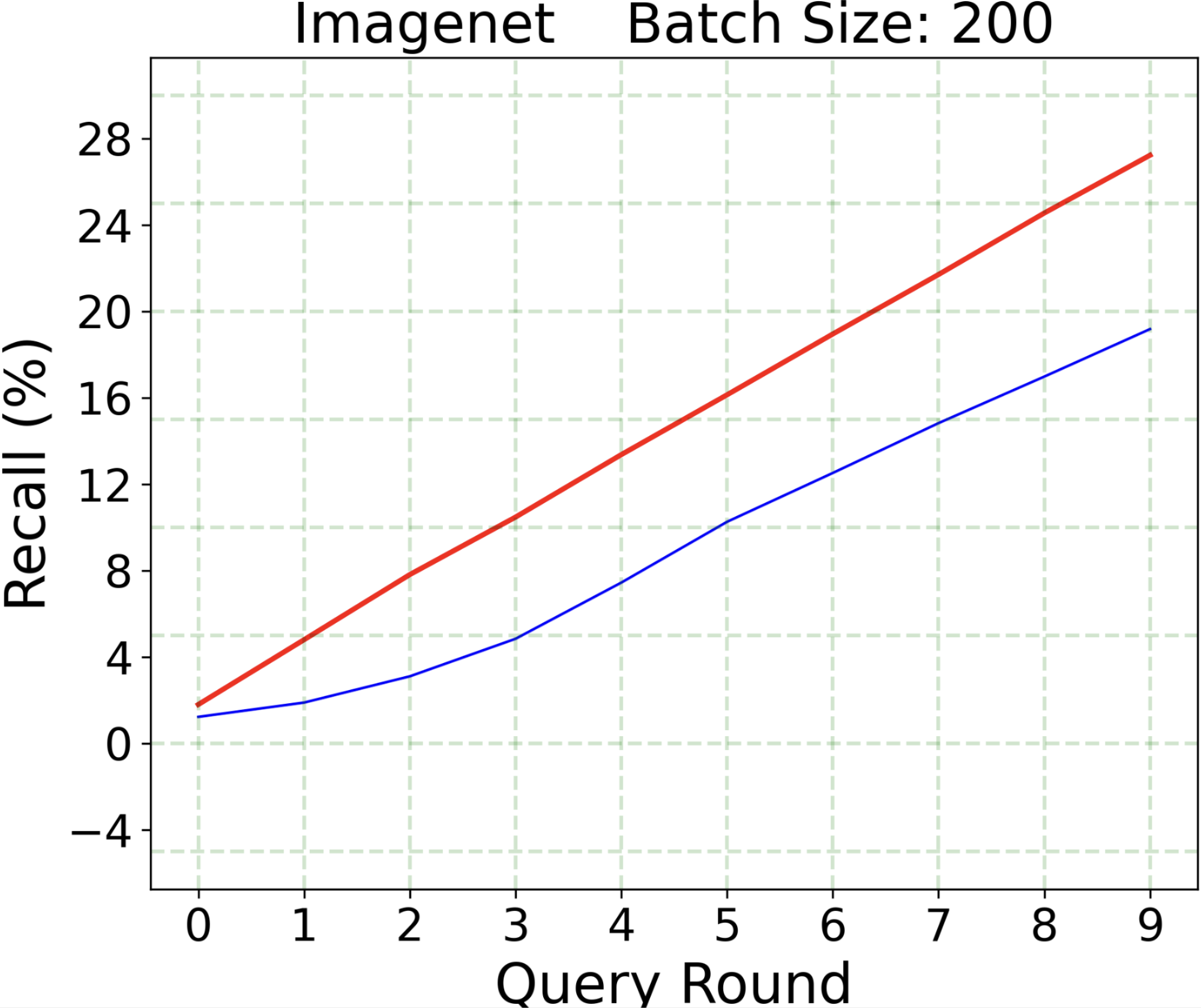} \\
    a) Accuracy &     b) Recall  \\
\end{tabular}
\begin{minipage}{\textwidth}
    \caption{We use Vision Transformer to train \textsc{Neat}, Red for the proposed method \textsc{Neat} and Blue for \textsc{LFOSA}.}
    \label{fig:vit}
\end{minipage}
\end{figure}

\begin{figure}[h]
    \centering
    \includegraphics[width=2\linewidth]{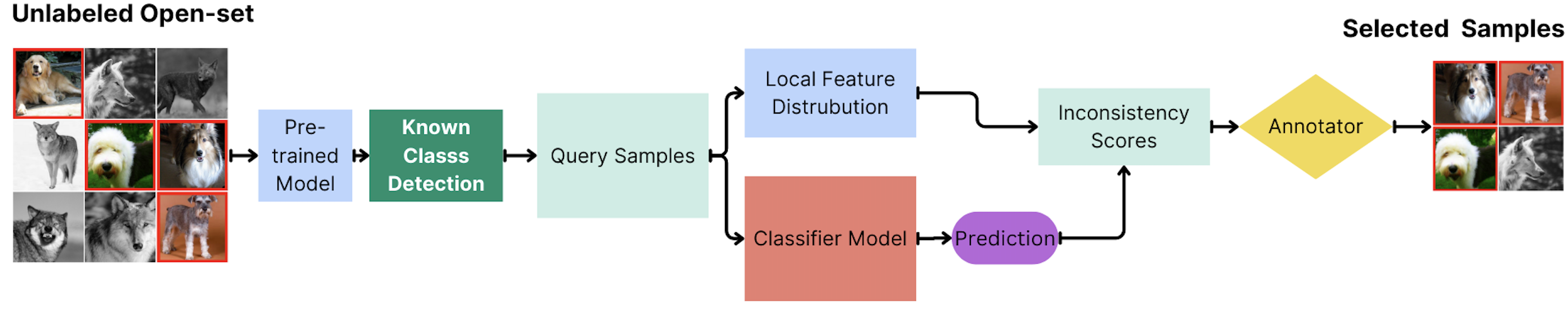}
    \begin{minipage}{\textwidth}
        \caption{The architecture of \textsc{Neat} includes a classifier model and a local feature distribution extractor, designed to identify both known and unknown samples.}
        \label{fig:enter}
    \end{minipage}
\end{figure}

We also employed ResNet-34, ResNet-50, and VGG-16 for training NEAT. However, ResNet-18 outperformed the others in terms of classification accuracy. Moreover, ResNet-18 required the least computational power among these models, making it a more efficient choice. The outcomes can be observed in Figures \ref{fig:structure}. In the subsequent experiment beside ResNet famliy we conducted experiments using the base Vision
Transformer (ViT) and present ImageNet results in Figure \ref{fig:vit}. The framework of NEAT can be viewed in Figures \ref{fig:enter}.

\clearpage

\begin{figure*}[p]
    \centering
    \setlength{\tabcolsep}{2pt}
    \begin{tabular}{ccc}
        \includegraphics[width=0.25\textwidth]{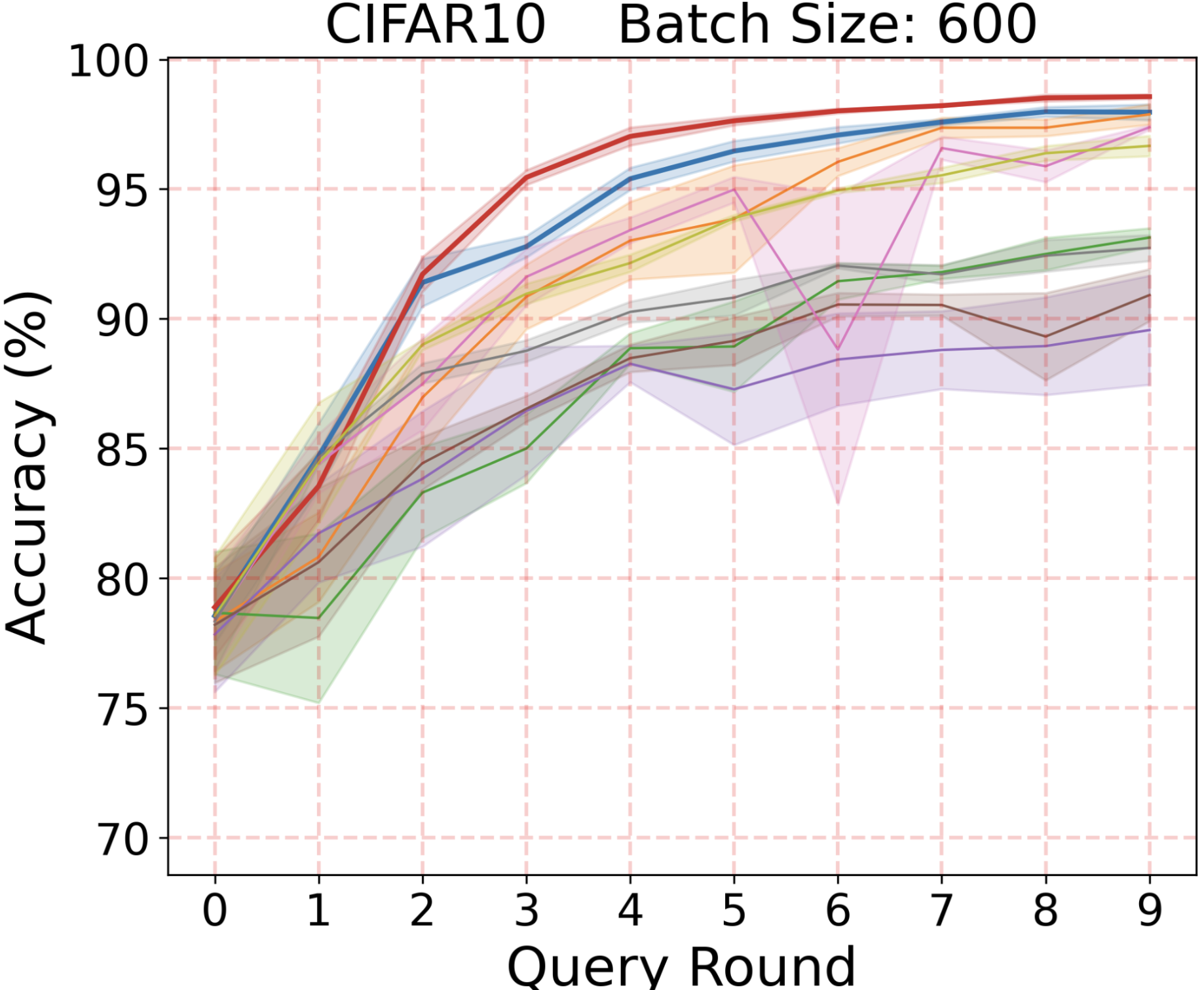} &
        \includegraphics[width=0.25\textwidth]{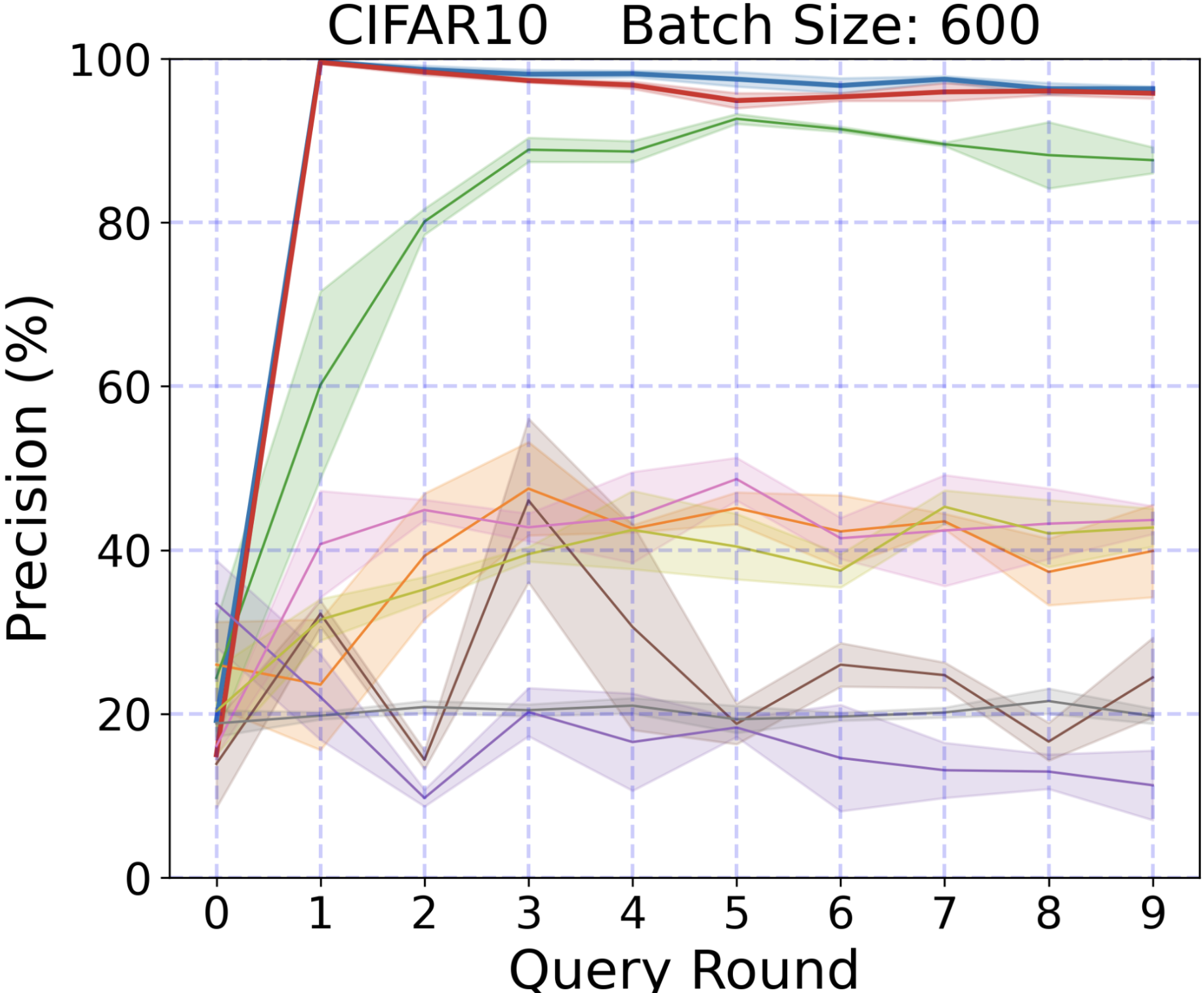} &
        \includegraphics[width=0.25\textwidth]{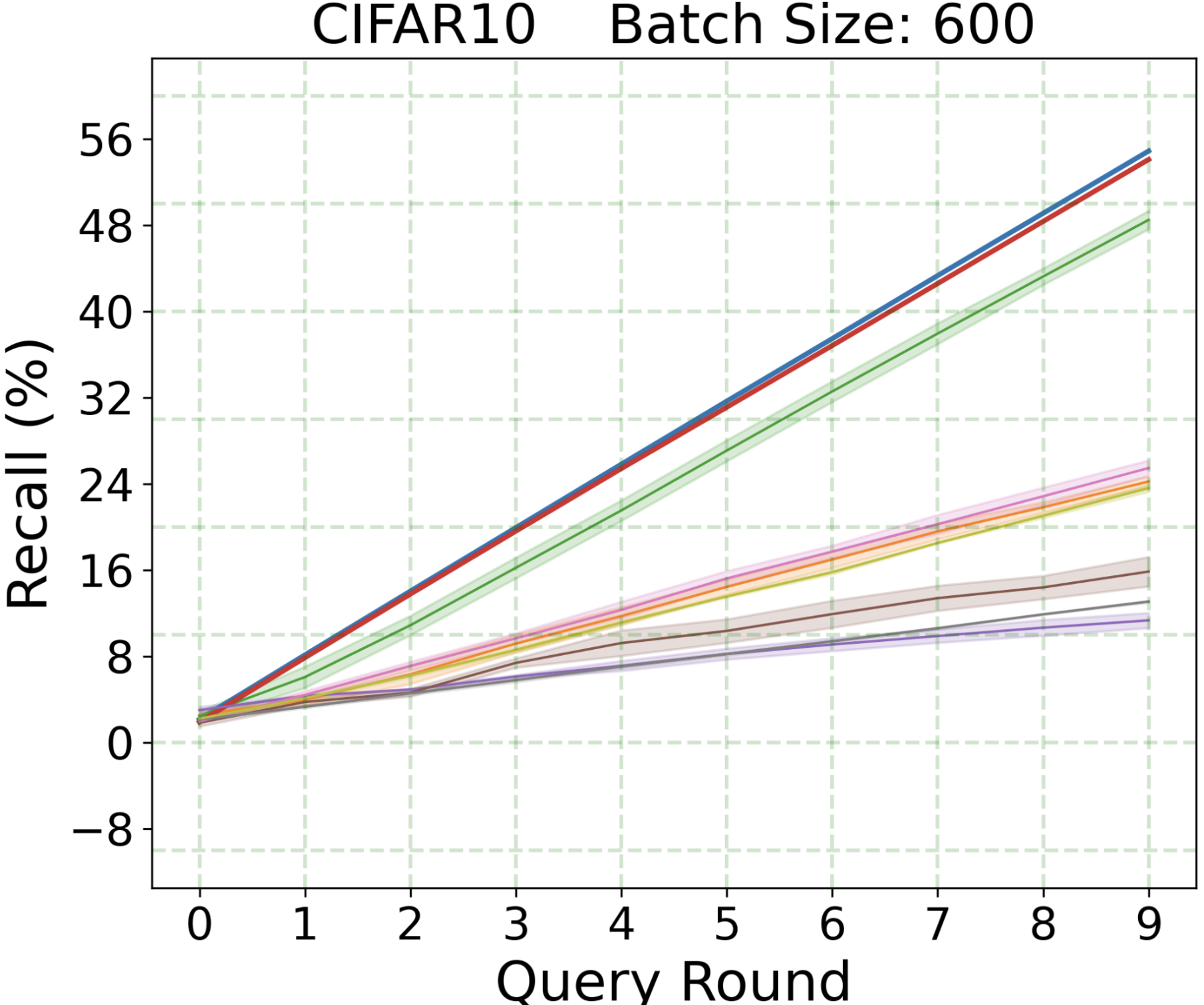} \\
        \includegraphics[width=0.25\textwidth]{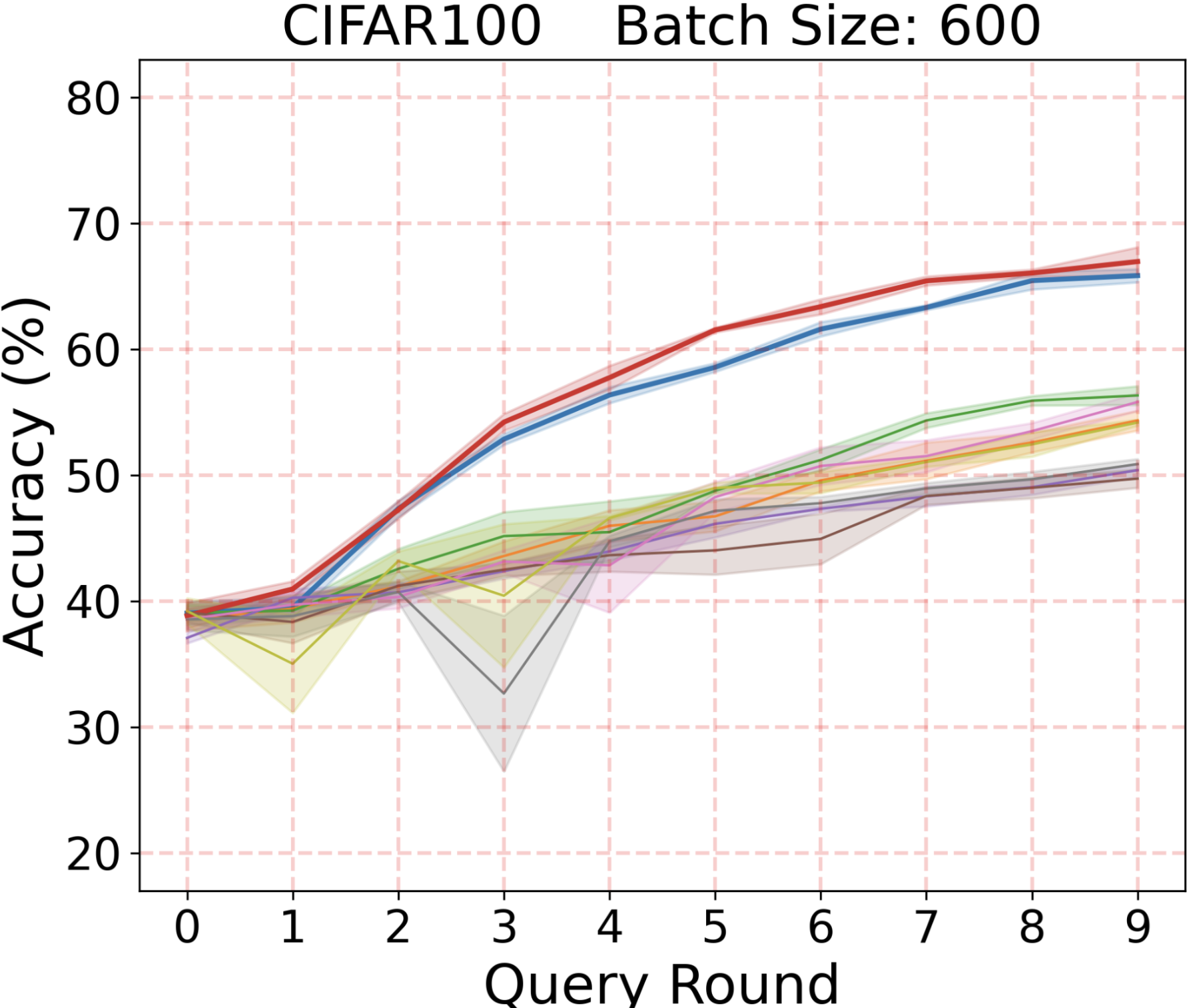} &
        \includegraphics[width=0.25\textwidth]{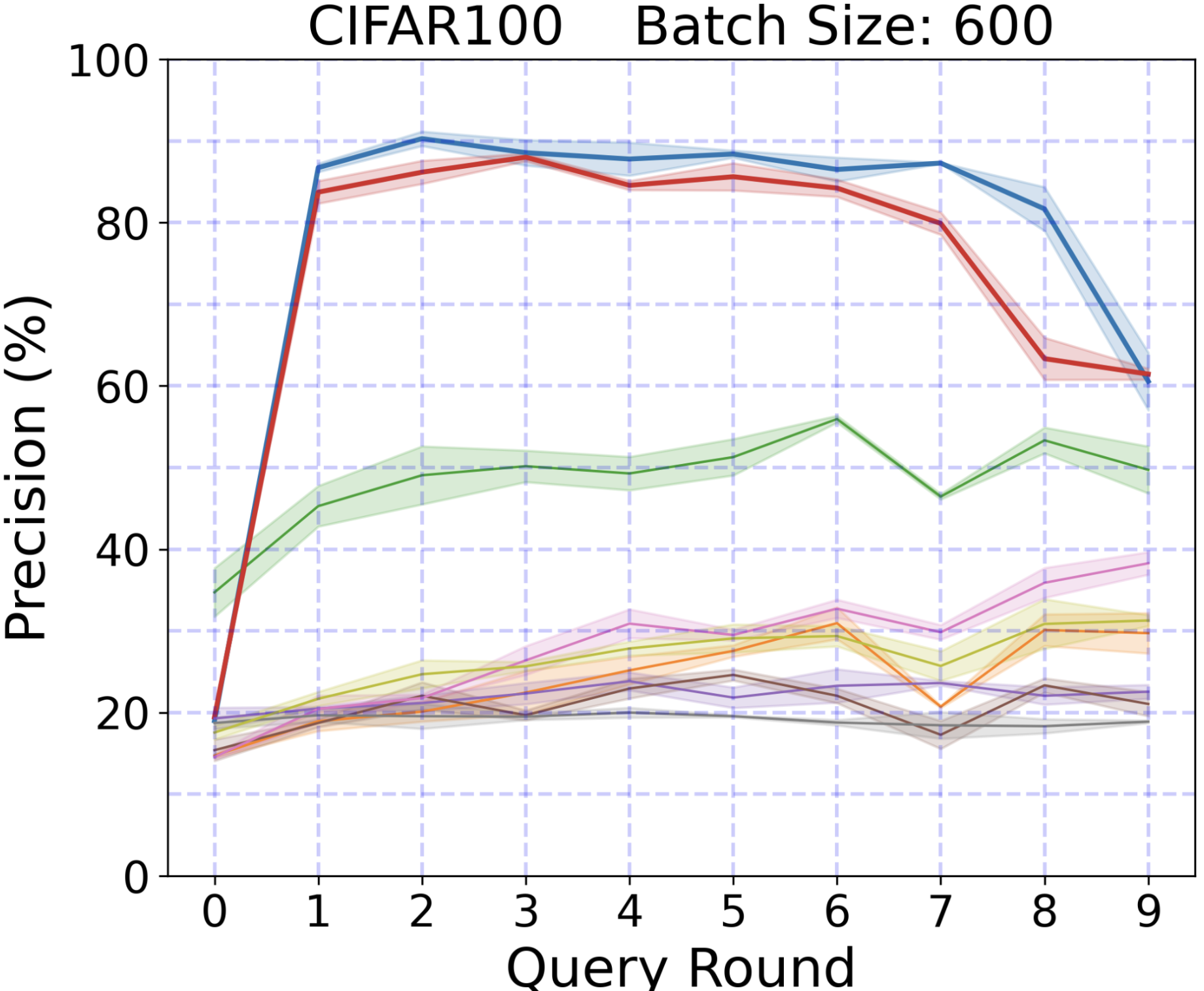} &
        \includegraphics[width=0.25\textwidth]{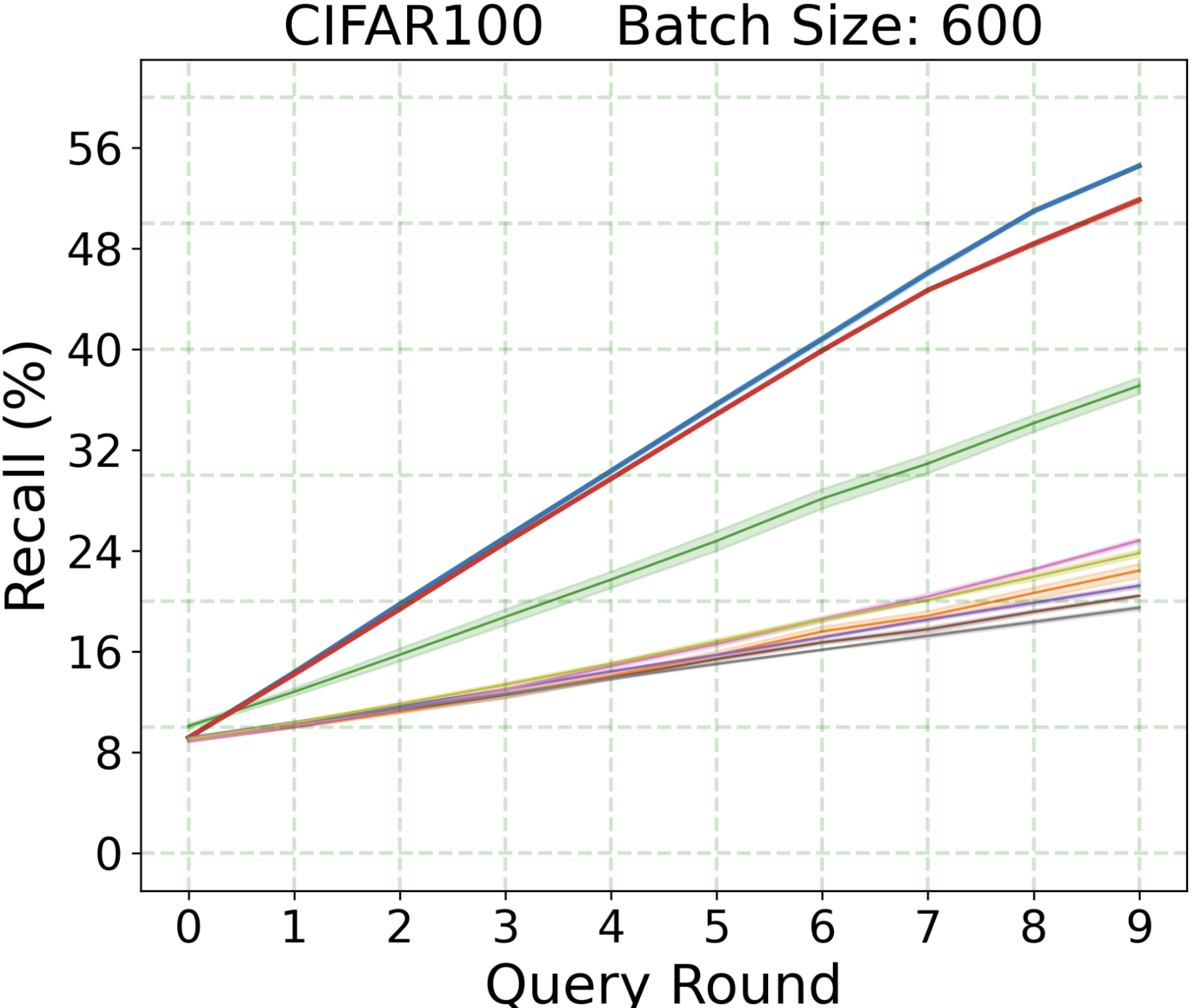} \\    
        \includegraphics[width=0.25\textwidth]{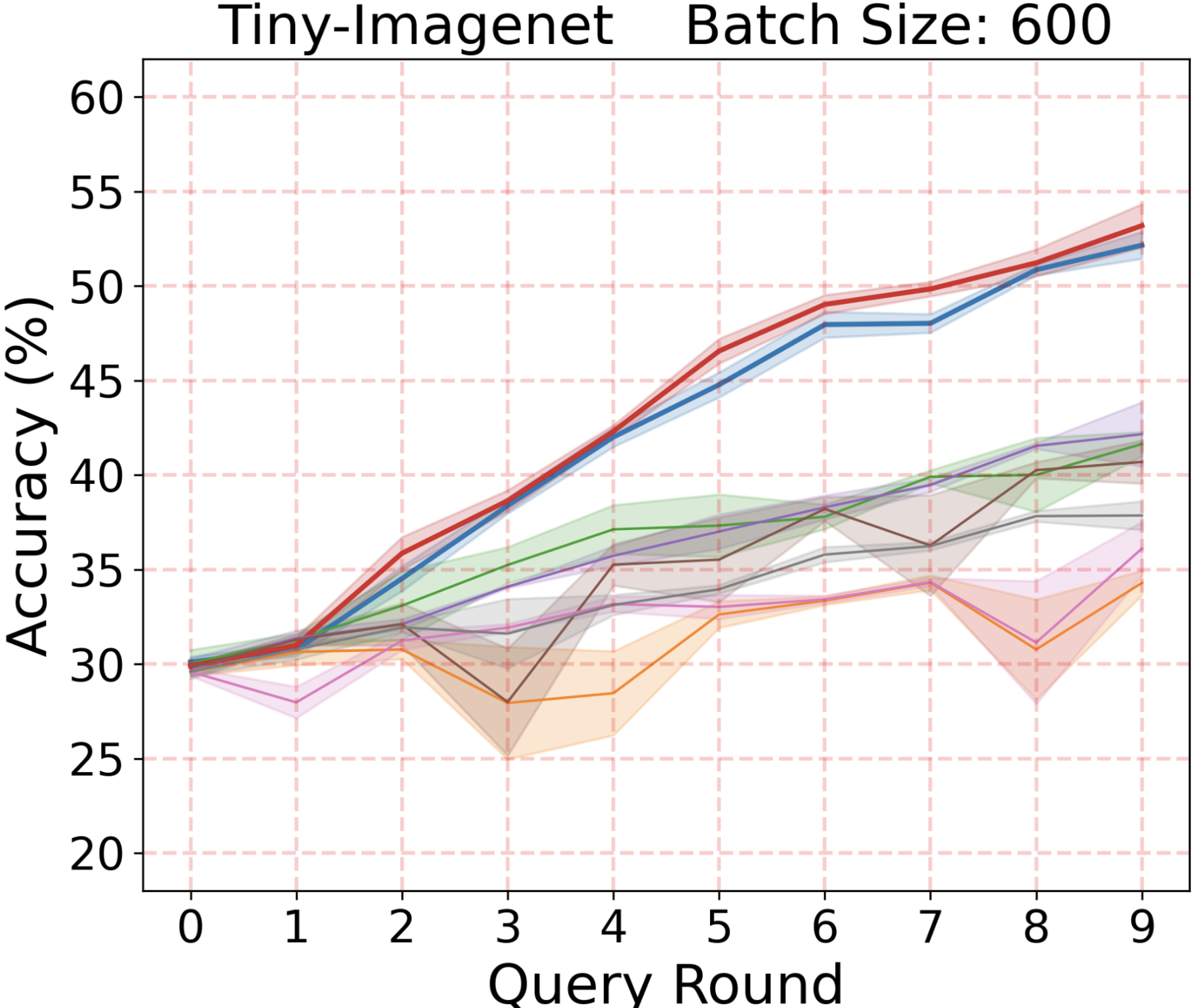} &
        \includegraphics[width=0.25\textwidth]{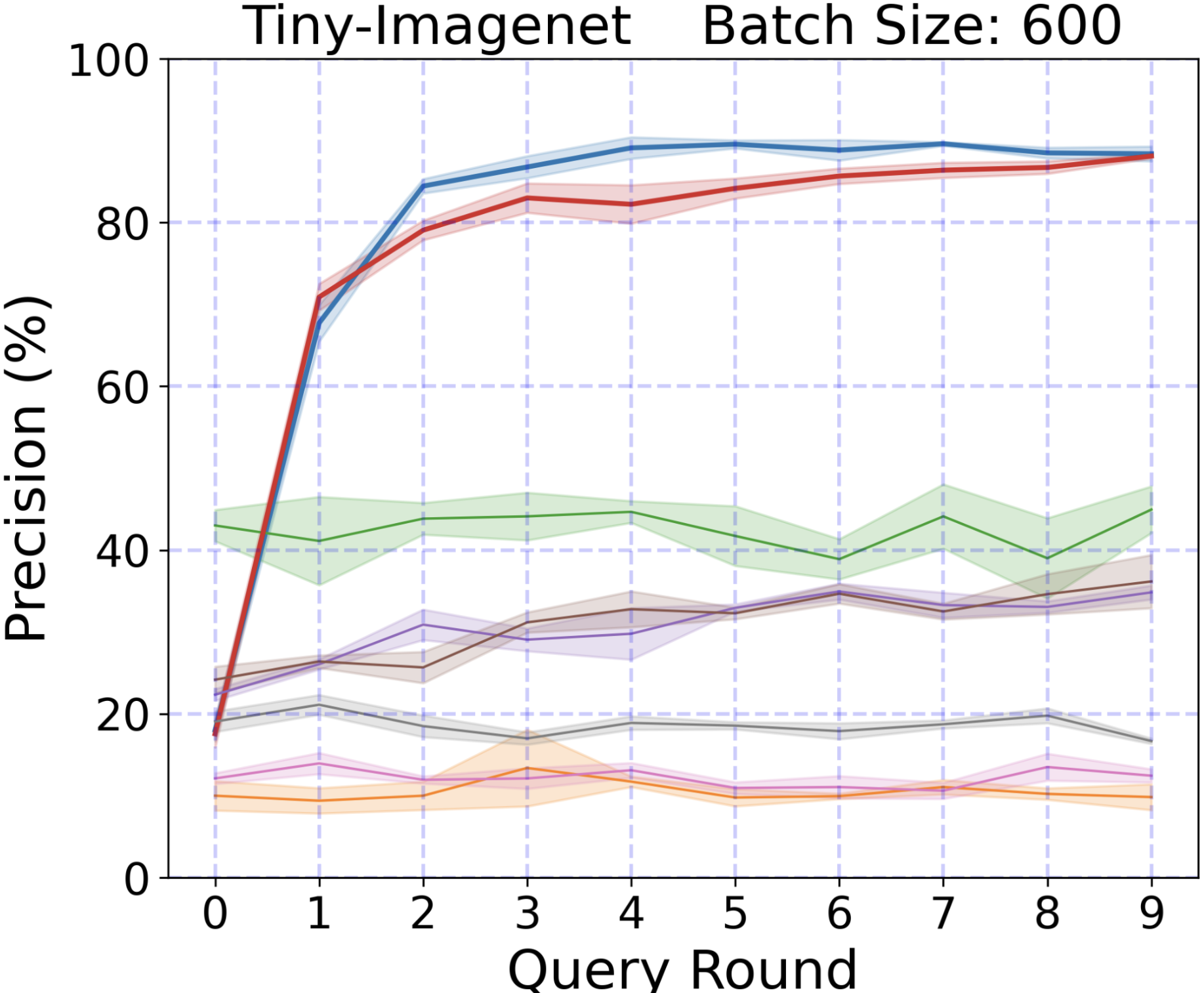} &
        \includegraphics[width=0.25\textwidth]{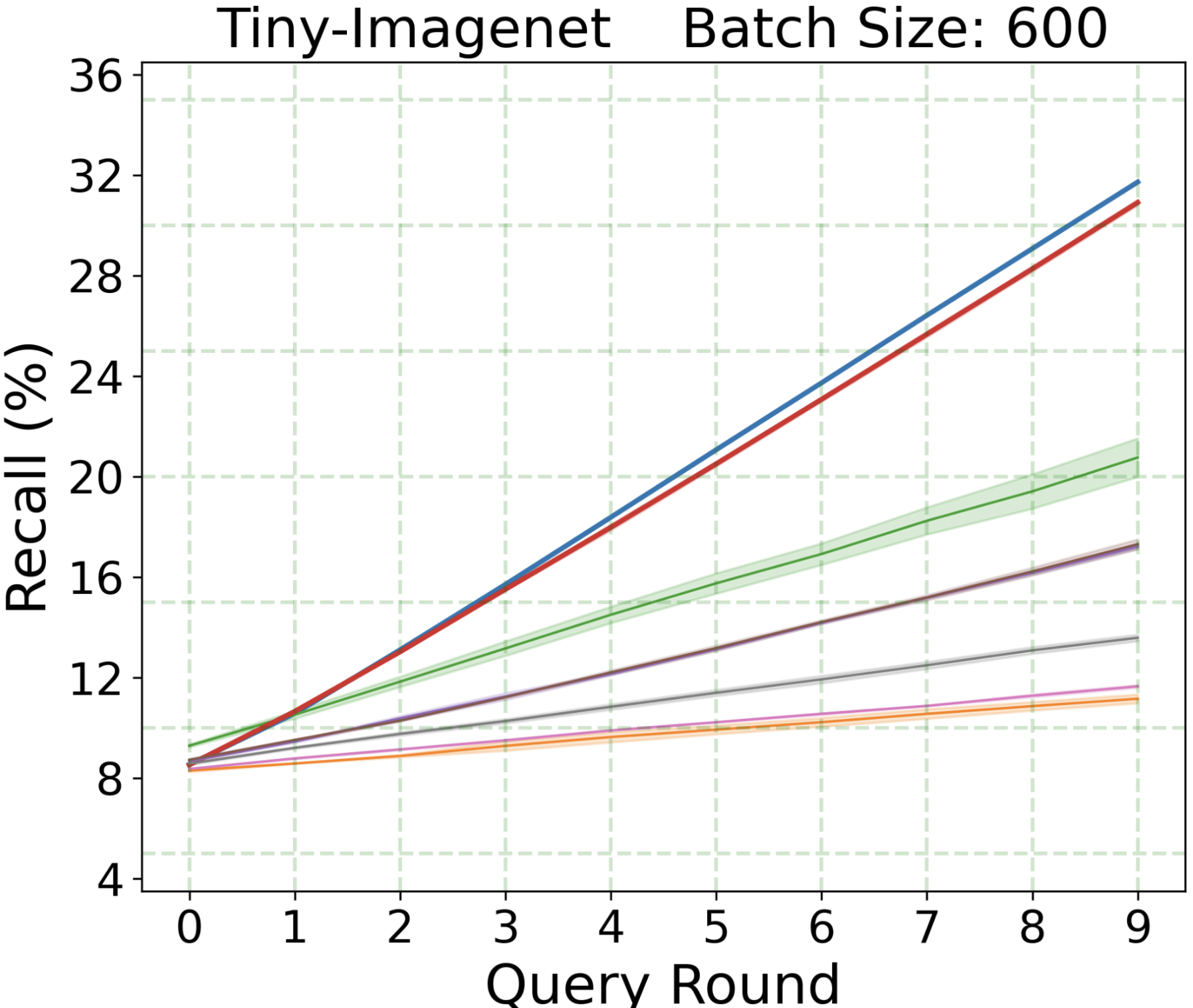} \\
        \multicolumn{3}{c}{\includegraphics[width=0.7\textwidth]{baseline/legund.png}} \\              
    \end{tabular}
     \caption{\textsc{Neat} achieves higher precision, recall and accuracy compared with existing active learning methods for active open-set annotation. We evaluate \textsc{Neat} and the baseline active learning methods on \textbf{CIFAR10}, \textbf{CIFAR100} and \textbf{Tiny-ImageNet} based on accuracy, precision and recall. }
    \label{fig:baseline}
\end{figure*}
\clearpage

\end{document}